\documentclass[lettersize,journal]{IEEEtran}
\usepackage{amsmath,amsfonts}
\usepackage{algorithmic}
\usepackage{array}
\usepackage[caption=false,font=normalsize,labelfont=sf,textfont=sf]{subfig}
\usepackage{textcomp}
\usepackage{stfloats}
\usepackage{url}
\usepackage{verbatim}
\usepackage{graphicx}
\def\BibTeX{{\rm B\kern-.05em{\sc i\kern-.025em b}\kern-.08em
    T\kern-.1667em\lower.7ex\hbox{E}\kern-.125emX}}
\usepackage{balance}
\usepackage{color}
\usepackage{dsfont}
\usepackage{graphicx}
\usepackage{svg}
\usepackage{booktabs}
\usepackage{rotating}
\usepackage{multirow} 
\usepackage{pifont}
\usepackage{hyperref}
\usepackage[table]{xcolor}

\begin{document}

\title{MuSc-V2: Zero-Shot Multimodal Industrial Anomaly Classification and Segmentation with Mutual Scoring of Unlabeled Samples}

\author{\IEEEauthorblockN{Xurui~Li,
Feng~Xue,~\IEEEmembership{Member,~IEEE},
and~Yu~Zhou~\IEEEmembership{Member,~IEEE}.}
\thanks{
This work was supported by the National Natural Science Foundation of China under Grant No.62176098. The computation is completed in the HPC Platform of Huazhong University of Science and Technology.
(Corresponding author: Yu Zhou.)

Xurui Li and Yu Zhou are with the School of Electronic Information and Communications, Huazhong University of Science and Technology, Wuhan 430074, China.
(e-mail: xrli\_plus@hust.edu.cn; yuzhou@hust.edu.cn).

Feng Xue is with the School of Computer Science,
University of Trento, Italy.
(e-mail: feng.xue@unitn.it).
}}

\maketitle

\begin{abstract}
Zero-shot anomaly classification (AC) and segmentation (AS) methods aim to identify and outline defects without using any labeled samples.
In this paper, we reveal a key property that is overlooked by existing methods:
normal image patches across industrial products typically find many other similar patches,
not only in 2D appearance but also in 3D shapes,
while anomalies remain diverse and isolated.
To explicitly leverage this discriminative property,
we propose a Mutual Scoring framework (MuSc-V2) for zero-shot AC/AS, which flexibly supports single 2D/3D or multimodality.
Specifically, 
our method begins by improving 3D representation through Iterative Point Grouping (IPG),
which reduces false positives from discontinuous surfaces.
Then we use Similarity Neighborhood Aggregation with Multi-Degrees (SNAMD) to fuse 2D/3D neighborhood cues into more discriminative multi-scale patch features for mutual scoring. 
The core comprises a Mutual Scoring Mechanism (MSM) that lets samples within each modality to assign score 
to each other, and Cross-modal Anomaly Enhancement (CAE) that fuses 2D and 3D scores to recover modality-specific missing anomalies. 
Finally, Re-scoring with Constrained Neighborhood (RsCon) suppresses false classification based on similarity to more representative samples.
Our framework flexibly works on both the full dataset and smaller subsets with consistently robust performance,
ensuring seamless adaptability across diverse product lines.
In aid of the novel framework,
MuSc-V2 achieves significant performance improvements:
a $\textbf{+23.7\%}$ AP gain on the MVTec 3D-AD dataset and a $\textbf{+19.3\%}$ boost on the Eyecandies dataset, surpassing previous zero-shot benchmarks and even outperforming most few-shot methods.
The code will be available at \href{https://github.com/HUST-SLOW/MuSc-V2}{GitHub}.
\end{abstract}


\begin{figure}[!t]
\centering
\includegraphics[width=0.49\textwidth]{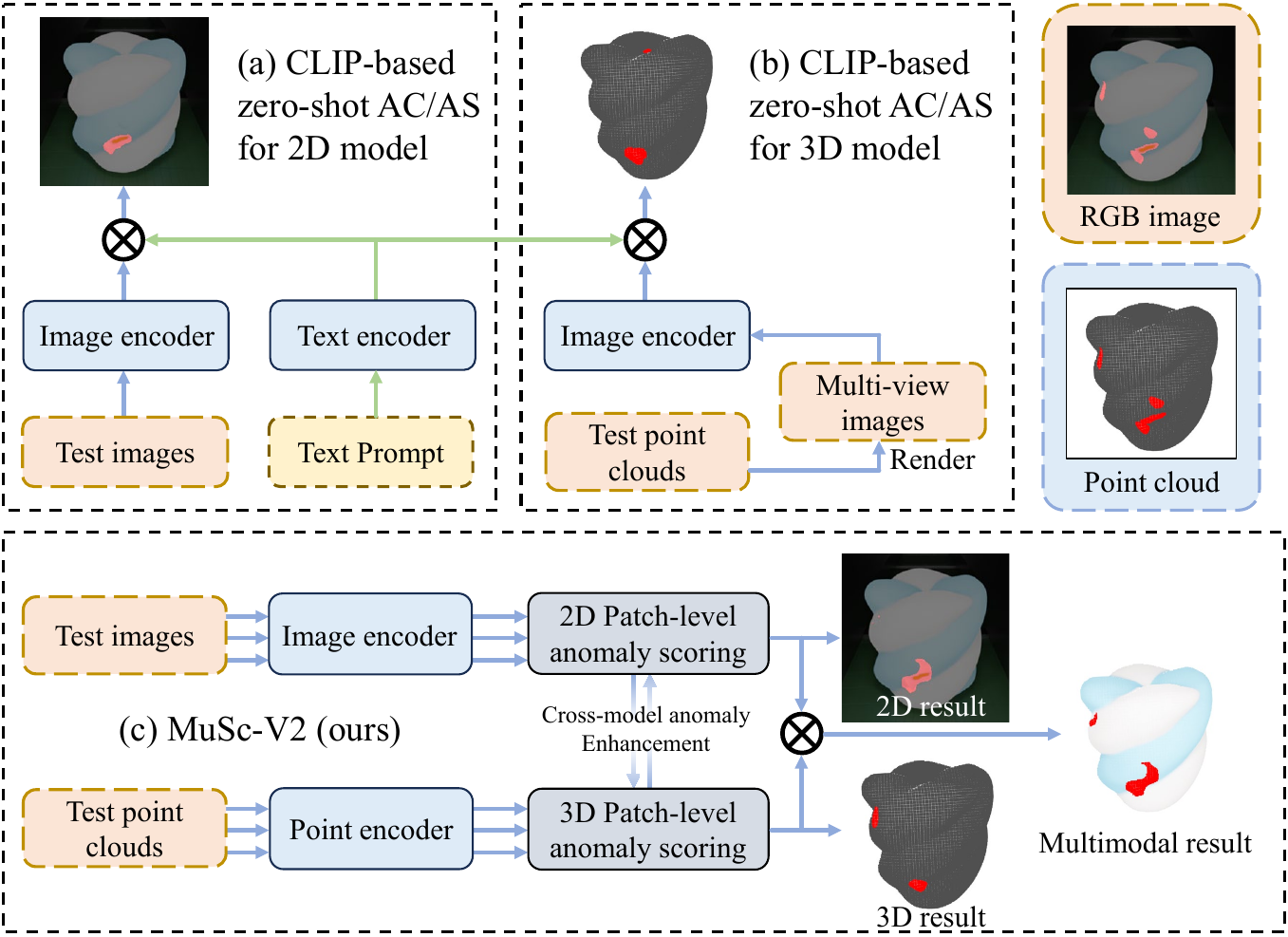}
\caption{
(a) Zero-shot AC/AS methods for 2D modal.
(b) Zero-shot AC/AS methods for 3D modal.
These CLIP-based methods require additional text prompts and fine-tuning on additional industrial datasets.
(c) Our MuSc-V2 is the first multimodal zero-shot method without any prompts or training.
}
\label{fig:intro}
\end{figure}

\section{Introduction}
\IEEEPARstart{I}{ndustrial} anomaly classification (AC) and segmentation (AS) are important tasks in the computer vision field.
The AC task aims to discover objects with anomalies at the sample level, while the AS locates anomalies precisely.
In real industrial scenarios, anomalies may appear on various objects, textures, shapes, and lights.
The high diversity of anomalies makes the AC/AS task challenging.
Current approaches address these challenges using 2D, 3D, or multimodal data.

Early industrial AC/AS searches focused on 2D images and evolved along three paradigms.
\textbf{Unsupervised approaches} (full-shot) achieve high accuracy but require a lot of labeled normal samples for training, e.g., \cite{eccv2024glass, tii2023cdo, pami2025diffusionad, tii2025cras, pami2023ssmctb, tip2cpr,pami2024NeuTraLAD, tii2024pnpt}.
In contrast, some \textbf{few-shot methods} \cite{ICLR2023graphcore,tii2023scosda,pami2021fewshotda,cvpr2024inctrl,cvpr2024promptad,cvpr2024mvfa} reduce this dependency, delivering competitive performance with minimal normal samples.
The latest \textbf{zero-shot approaches} \cite{CVPR2023winclip,arxiv2023APRILGAN,nips2023ACR} shown in Fig. \ref{fig:intro} (a) break new ground by using CLIP's text-image alignment for anomaly measurement.

To overcome the limitations of 2D imaging,
such as illumination, angle occlusion, camera resolution, etc.,
recent methods \cite{cvpr2023m3dm,cvpr2023btf, pami2025m3dm-nr} have introduced the 3D point cloud.
These approaches complement ambiguous 2D features with spatial information,
culminating in zero-shot 3D techniques \cite{nips2024pointad} that employ multi-view rendering,
as shown in Fig. \ref{fig:intro} (b).

However, existing methods, whether 2D or 3D, rely on comparing each unlabeled sample with labeled normal samples or text prompts.
This convention overlooks the wealth of implicit normal information among unlabeled samples across both 2D and 3D modals.
This is particularly evident in industrial production lines, where products from the same line exhibit strong consistency and homogeneity.
Actually, our statistics reveals that normal primitives (i.e., 2D pixels or 3D points) dominate the industrial data (${99.71\%}$ in MVTec 3D-AD, ${99.77\%}$ in Eyecandies 2D images; ${98.57\%}$-${99.31\%}$ 3D points).
This prevalence allows normal regions to consistently find numerous similar counterparts across other unlabeled samples.
In contrast, anomalies are often different from each other even with the same type,
due to their randomness and unpredictability.
Therefore, such an intrinsic discriminative property enables exploitation of both normal consistency and abnormal divergence in unlabeled multimodal data for zero-shot AC/AS.

Motivated by this, we propose MuSc-V2, a multimodal zero-shot AC/AS approach (Fig. \ref{fig:intro} (c)),
which directly recognizes anomalies by scoring it using other unlabeled samples mutually,
thus no need for any training process or prompts.   
To enable effective mutual scoring, we first develop two feature improvement modules critical for reducing false detections.
For 3D modal, Iterative Point Grouping (IPG) replaces traditional KNN grouping to reduce false positives caused by discontinuous surfaces, ensuring geometrically consistent 3D patches.
For both modalities, Similar Neighborhood Aggregation with Multi-Degrees (SNAMD) models variable-sized anomalies by fusing multi-scale neighborhood features.
During aggregation, we design similarity-weighted pooling (SWPooling) to prevent missed detections.
Building upon the unexploited discriminative characteristic implied in the unlabeled data, our core Mutual Scoring Mechanism (MSM) leverages the improved 2D/3D features to establish a training-free paradigm where unlabeled samples mutually assign anomaly scores.
Meanwhile, to address modality-specific false negatives, we propose the Cross-modal Anomaly Enhancement (CAE) to raise scores of unnoticeable abnormal regions in a single modality.
Finally, to suppress false classification caused by local noise and weak anomalies, we explore the sample-level relationship and further design the Re-Scoring with Constrained Neighborhood (RsCon).
Evaluations on MVTec 3D-AD and Eyecandies datasets demonstrate significant improvements in both single-model (2D or 3D) and multi-modal (2D+3D).
Especially in the multimodal AS, we obtain $\textbf{+23.7\%}$ and $\textbf{+19.3\%}$ gains on MVTec 3D-AD and Eyecandies datasets.
For the multimodal AC, our method achieves $\textbf{+6.2\%}$ and $+1.2\%$ AUROC on these datasets.
It is worth noting that these performances remain consistent across the entire dataset and smaller subsets (drop $\leq$ 1.0\%).
Moreover, our method shows strong robustness to varying ratios of normal samples, even in the extreme case with no normal sample available, performance degradation remains below 3\%.
These results further attest to its reliability for real-world scenarios.

The points below highlight the contributions of our work:
\begin{itemize}
\item To the best of our knowledge,
we propose the first multimodal method MuSc-V2 that only uses the unlabeled samples for industrial AC and AS. 
Furthermore, our method is also the first training-free multimodal (2D and 3D) zero-shot industrial AC/AS method.
\item We reveal the potential capability of normal and abnormal regions contained in unlabeled samples regardless of 2D or 3D modal.
It inspires us to propose the mutual scoring mechanism, which is a novel zero-shot AC/AS paradigm with high flexibility and adaptability to any modal.
\item Our method has significant advantages compared with the existing zero-shot methods, and such advantages are applicable in 2D, 3D, and 2D+3D settings.

\end{itemize}

This paper is an extension of our previous version, i.e. MuSc published in ICLR-2024 \cite{iclr2024musc}.
Compared with the conference version, the following improvements have been made:
\begin{itemize}
\item[1) ] \textbf{Framework extension.}
We extend the original 2D framework to multimodal (2D+3D) and conduct experiments on multimodal datasets to verify its effectiveness.
\item[2) ] \textbf{Module improvement.}
During the point cloud pre-processing, we replace the traditional KNN with our Iterative Point Grouping (IPG) to generate consistent normal features.
For fine-grained anomaly segmentation, we update the previous LNAMD to SNAMD, and adjust it to adapt to the 3D modal.
The SWPooling is designed to reduce the missed detections.
If both modals are available, we incorporate the Cross-modal Anomaly Enhancement (CAE) into the vanilla MSM to boost detection of modality-specific anomalies.
We improve the original RsCIN to RsCon, which could be compatible with 3D backbones without the $\texttt{[CLS]}$ token.

\item[3) ] \textbf{Performance gain.}
Experiments demonstrate that MuSc-V2 is $\mathbf{5.6}\times$ faster than the original MuSc.
Furthermore, MuSc-V2 achieves $\mathbf{+0.8\%}$ AP segmentation on MVTec AD and $\mathbf{+1.6\%}$ F1-max improvement on VisA.

\item[4) ] \textbf{More scenarios.}
We investigate the robustness of MuSc-V2 on datasets with different sizes and the ratios of normal samples, which proves that MuSc-V2 could generalize to different production lines without any training.

\end{itemize}

\section{Related Works}
\subsection{Transformer architecture for 2D and 3D representation}
Vision transformer (ViT) \cite{ICLR2021ViT} and point transformer (PT) \cite{ICCV2021pointTransformer} have become standard for 2D and 3D feature representation.
Some pre-trained models like CLIP \cite{ICML2021CLIP}/DINO \cite{iccv2021dino} for 2D modal and Point-MAE \cite{eccv2022pointmae}/Point-BERT \cite{cvpr2022pointbert} for 3D modal deliver high-quality patch features.
However, these features often struggle with industrial anomalies of varying sizes.
Swin transformer \cite{iccv2021swin,pami2022vlt} proposes the varied-size window attention to compute attention within multi-scale windows, which risks compromising fine-grained anomaly discrimination.
We propose a training-free solution SNAMD, that optimizes patch features via Similarity-Weighted Pooling to better capture multi-scale anomalies and preserve small anomalies.

\subsection{Point cloud grouping}
To reduce computational costs of self-attention, existing 3D feature extractors \cite{cvpr2017pointnet, tpami2023Flatteningnet, ICCV2021pointTransformer, pamivote2cap,cvm2021pct, pami2024genvcl} preprocess point clouds through FPS and KNN grouping,
encoding each group as a 3D patch token.
However, these methods risk merging multiple surfaces within a single group, particularly when components are spatially proximate, resulting in deviant tokens that trigger false positives.
Point Transformer V2/V3 \cite{nips2022pointtranformerv2, cvpr2024pointtranformerv3} propose new grouping strategies, but they focus more on optimizing speed and overlook this fundamental issue.
We propose an Iterative Point Grouping strategy to address this challenge by ensuring surface-consistent groupings for more robust 3D feature extraction.

\begin{figure*}[t]
\vspace{-.1in}
\begin{center}
\includegraphics[width=1\textwidth]{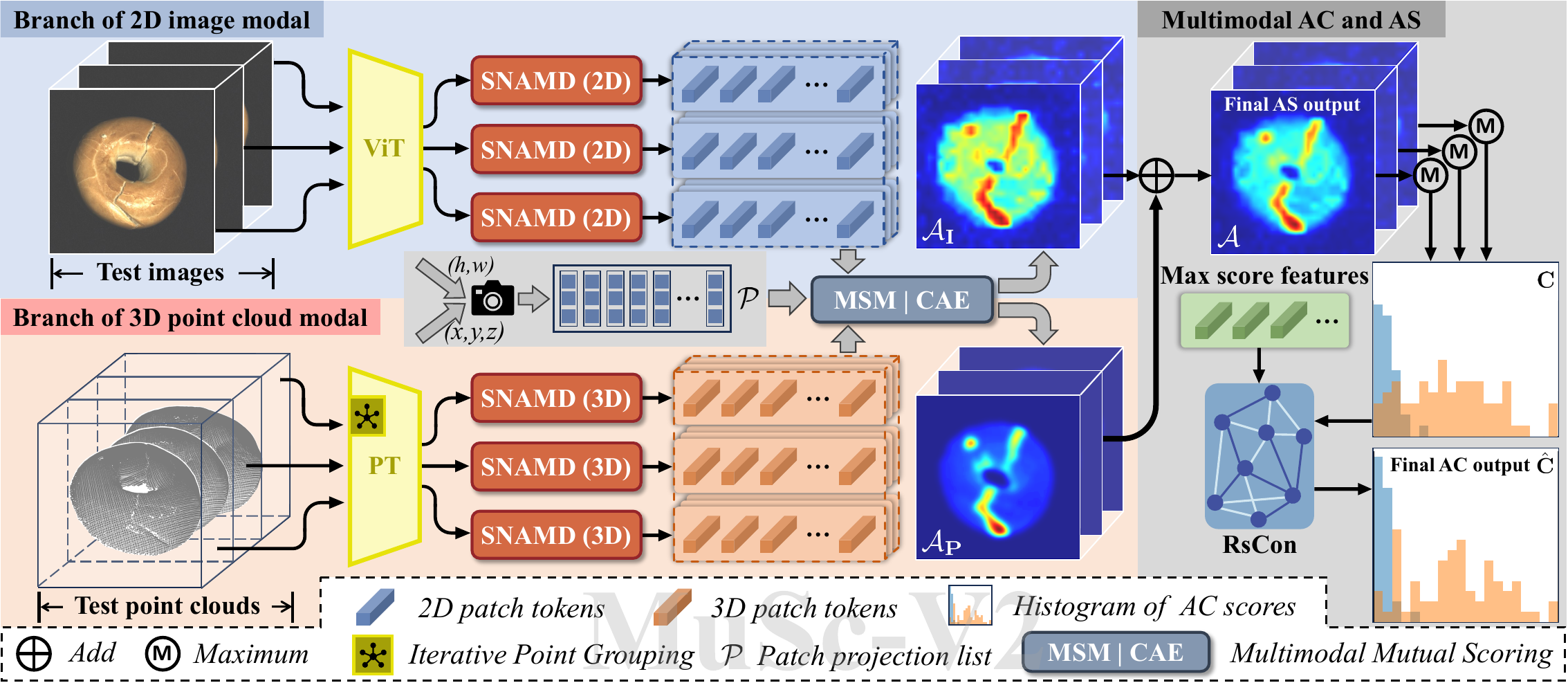}
\vspace{-5mm} 
\caption{
\textbf{The pipeline of our MuSc-V2.}
This framework processes 2D images and 3D point clouds through four important innovations:
(1) IPG replaces the current grouping strategy in the point transformer to generate groups with continuous surfaces (Sec. \ref{sec:2d_feature}).
(2) SNAMD improves the abnormal modeling ability with varying sizes for both modals (Sec. \ref{sec:3d_feature}).
(3) MSM obtains anomaly segmentation results of 2D/3D modals. CAE enhances scores of anomalies if both modals are available (Sec. \ref{sec:m3sm}).
(4) RsCon reduces false anomaly classification from local noise and weak anomalies (Sec. \ref{sec:rscon}).
}
\label{fig:pipeline}
\end{center}
\vspace{-10pt}
\end{figure*}

\subsection{Zero-shot anomaly classification and segmentation}
In the industrial vision field, zero-shot AC/AS has garnered more attention.
However, most methods focus on the 2D modal by image-text alignment of CLIP model \cite{ICML2021CLIP}.
These CLIP-based approaches \cite{CVPR2023winclip, eccv2024adaclip, eccv2024vcpclip, wacv2024promptad, arxiv2023APRILGAN, mm2024filo} fine-tune image encoder or text encoder to bridge the domain gap between the natural and industrial scenarios.
Meanwhile, some zero-shot methods \cite{wacv2023utad, nips2023ACR, iclr2024musc} focus on detecting anomalies by using the unlabeled samples itself.
\cite{wacv2023utad} explores the relationship between patches inside one unlabeled image, but only handles texture products.
ACR \cite{nips2023ACR} proposes a new adaptation strategy without human involvement, which trains the network by other products from the dataset.
For the 3D zero-shot task, PointAD \cite{nips2024pointad} renders the point cloud to multiple images with different view angles.
In this way, 2D methods could be used to process 3D data.

Unlike general zero‑shot learning \cite{cvpr2025vspcn, iccv2025rare} (non‑industrial one), where target domain data often contain diverse categories with large intra‑sample variations, industrial zero‑shot learning benefits from the inherent consistency of products manufactured on the same production line.
This consistency means that unlabeled industrial samples share strong local similarities with each other.
However, existing industrial zero‑shot methods typically overlook the rich relational information among unlabeled samples.
Inspired by this observation, we propose a mutual scoring mechanism for both 2D/3D modals, which explicitly exploits the inter‑sample relationships within industrial unlabeled datasets without additional fine-tuning.

\subsection{Multimodal anomaly classification and segmentation}
Multimodal industrial AC/AS task aims to identify defects in the industrial product through its 2D image and 3D point cloud.
Some methods \cite{eccv2025r3d,cvpr2024anomaly-shapenet,mm2023easynet,cvpr2023btf} are proposed for the unsupervised (full-shot) AC/AS task.
Among them, M3DM \cite{cvpr2023m3dm} fine-tunes features of two modals by contrastive learning for cross-modal alignment.
Shape-guided \cite{icml2023shape-guided} use 2D features stored in the memory bank to reconstruct 3D features to guide the identification of 2D anomalies.
In addition, CFM \cite{cvpr2024cfm} implements cross-modal feature reconstruction according to the student-teacher network, greatly reducing time consumption.
These methods require collecting a large number of normal images and point clouds for model training,
which limits the migration ability on the new production line.
We propose the multimodal mutual scoring without any training and labeled samples.
The CAE module is inserted into our mutual scoring mechanism to eliminate the blind spot of single model.

\subsection{Manifold learning}
In high‑dimensional feature spaces, the data often lie on a low‑dimensional manifold.
Euclidean distances computed directly in the original space can be unreliable because they do not reflect the underlying manifold geometry, only local neighborhoods preserve Euclidean properties.
To address this, some manifold learning methods \cite{nips2003RDM,pami2008riemannian, CVPR2012sd,pami2011adaptml}  construct an embedding space where distances approximate geodesic distances along the manifold.
Inspired by the above principles, we develop the RsCon module to refine pixel-level anomaly classification by leveraging the local manifold structure.
Each sample's anomaly score is refined jointly with scores of its most similar neighbors on the manifold, ensuring that score calibration respects the intrinsic data geometry.

\section{Method}
\label{sec:method}
Given $N$ unlabeled image-point cloud pairs $\mathcal{D}\!=\!\{O_i | i\!\in\![1,N],O_i\!=\!(I_i, P_i)\}$, where $I_i$ and $P_i$ is the 2D image and 3D point cloud respectively.
Our approach is designed to handle both single-modal and multi-modal situations.
The pipeline is illustrated in Fig. \ref{fig:pipeline}, which consists of four important components:
(1) \textbf{2D/3D feature representation} (Sec. \ref{sec:2d_feature}).
We extract the pretrained 2D/3D features in this section.
For 3D model, we design Iterative Point Grouping (IPG) strategy to replace the traditional pre-processing method, KNN, to alleviate 3D false positives from discontinuous surface representations.
(2) \textbf{Similar Neighborhood Aggregation with Multiple Degrees} (Sec. \ref{sec:3d_feature}).
We propose SNAMD module to aggregate multi-scale neighborhood features to suppress false negatives of variable-sized anomalies.
(3) \textbf{Mutual Scoring Mechanism} (Sec. \ref{sec:m3sm}).
This zero-shot AC/AS paradigm scores 2D/3D patches in an unlabeled sample using other unlabeled samples.
Moreover, to reduce modality-specific false negatives, we propose the Cross-modal Anomaly Enhancement (CAE) to fuse 2D/3D scores.
(4) \textbf{Re-Scoring with Constrained Neighborhood} (Sec. \ref{sec:rscon}).
This module suppresses false classifications caused by local noise and weak anomalies.

\subsection{2D/3D Patch Representation}
\label{sec:2d_feature}

\textbf{2D Feature Extraction.} 
Following \cite{CVPR2023winclip,arxiv2023APRILGAN, ICLR2024anomalyclip}, we adopt a vision transformer \cite{ICLR2021ViT} consisting of $S$ stages to extract hierarchical 2D features.
For image $I_i$, we define the patch tokens produced by stage $s$ as $F_\textbf{I}^{i, s} \!\in\! \mathbb{R}^{M_\textbf{I} \times C_\textbf{I}}$, where $M_\textbf{I}$ is the number of patch tokens, 
$C_\textbf{I}$ is the feature dimension,
and $s\in[1,S]$.

\textbf{3D Feature Extraction.}
To extract patch-level 3D features, we first adopt an Iterative Point Grouping strategy (introduced below) as a pre-processing step to group points that lie on common surfaces.
Then, following \cite{cvpr2023m3dm, cvpr2023btf, cvpr2024cfm},
the point groups are fed into a point transformer \cite{ICCV2021pointTransformer} with $S$ stages to extract the 3D features.
For point cloud $P_i$, the stage $s$ produce $M_\textbf{P}$ 3D patch tokens as $F_\textbf{P}^{i, s} \!\in\! \mathbb{R}^{M_\textbf{P} \times C_\textbf{P}}$, where each group is regarded as a 3D patch.

\begin{figure}[!t]
\centering
\includegraphics[width=0.49\textwidth]{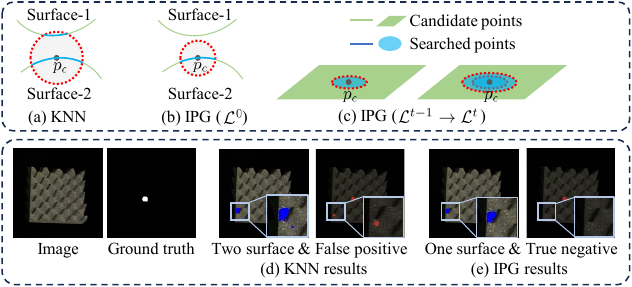}
\caption{
\textbf{Toy example of searching $K_\textbf{P}$ points for the center point $p_\textbf{c}$.}
The \textcolor[rgb]{0.66,0.82,0.56}{green lines and regions} represent the candidate points,
and the \textcolor[rgb]{0.42,0.56,0.82}{blue ones} indicate the searched points as the group points of $p_\textbf{c}$.}
\label{fig:point_group}
\end{figure}

\textbf{Iterative Point Grouping.}
Existing 3D feature extractors \cite{ICCV2021pointTransformer, cvpr2017pointnet, cvm2021pct} employ farthest point sampling \cite{TIP1997fps} and KNN strategy to cluster points for local feature extraction.
However, the KNN strategy employs spatial proximity only,
which may result in one group containing discontinuous surfaces, as shown in Fig.\ref{fig:point_group} (a).
Such discontinuous normal point groups are easily misclassified as anomalies due to their isolated pattern.
To alleviate this problem,
we propose the Iterative Point Grouping (IPG) strategy,
which replaces fixed-distance neighborhood selection with an iterative expansion approach.
We first group the point cloud $P_i$ into $M_\textbf{P}$ groups of $K_\textbf{P}$ points by KNN.
The center point of each group is represented as $p_\textbf{c}$.
The following steps are carried out on this basis above.

\begin{itemize}
\item[1. ]
\textit{Curvature Calculation:}
To correct point groups with discontinuous surfaces,
we compute the curvature $\mathcal{C}$ at each group's center point $p_\textbf{c}$ \cite{cdp2016differential}.
Specifically, we define the $K_\textbf{P}$ points nearest to $p_\textbf{c}$ as $\{p_1,...,p_{K_\textbf{P}}\}$.
Then we normalize the coordinates of $p_i$ as follows,
\begin{equation}
\hat{p}_i = \frac{\overline{p}_i}{\text{max}(\Vert \overline{p}_1 \Vert_2,...,\Vert \overline{p}_{K_\textbf{P}} \Vert_2)}
\end{equation}
where $\overline{p}_i = p_i - p_\textbf{c}$.
Furthermore, we calculate the eigenvalues $\{\lambda_1,\lambda_2,...\}$ of the covariance matrix of the points $\{\hat{p}_1,...,\hat{p}_{K_\textbf{P}}\}$.
The curvature $\mathcal{C}$ is calculated by,
\begin{equation}
\mathcal{C} = \frac{\text{min}(\lambda_1,\lambda_2,...)}{\text{sum}(\lambda_1,\lambda_2,...)+\epsilon}
\end{equation}
where $\epsilon$ is a small constant to prevent zero denominator.
As shown in Fig.\ref{fig:point_group} (a), we observe that when points from different surface (Surface-1) are incorrectly grouped, the curvature at $p_\textbf{c}$ increases significantly.
Therefore, we perform the following steps for re-grouping if $\mathcal{C}$ exceeds a predefined threshold $\mathcal{C}_{thr}$.

\item[2. ]
\textit{Group Initialization:}
To re-group the points around $p_\textbf{c}$ with over-threshold curvature,
we initialize a group $\mathcal{L}^0\!\!=\!\!\{p_j\}_{j=1}^{K_\text{iter}}$ containing $K_\text{iter}$ nearest points of $p_\textbf{c}$,
where $K_\text{iter} < K_\textbf{P}$ remains small enough to avoid including points from other planes.
As shown in Fig.\ref{fig:point_group} (b), these initial points identify a unique surface (\textcolor[rgb]{0.42,0.56,0.82}{blue line}) for each group,
effectively preventing the inclusion of points from other nearby surfaces (Surface-1).

\item[3. ]
\textit{Group Expansion:}
To expand the group $\mathcal{L}^0$ on a continuous surface,
we iteratively add the $K_\text{iter}$ points closest to this group to $\mathcal{L}^0$.
Specifically, for each candidate point $p_{\hat{j}} \in P_i$ in iteration $t$, we compute its distance $d(t,p_{\hat{j}})$ to the current group $\mathcal{L}^{t-1}$ as,
\begin{equation}
d(t,p_{\hat{j}}) = \underset{p_j\in \mathcal{L}^{t-1}}{\min}\Vert p_{\hat{j}} - p_j \Vert_2.
\end{equation}
Then we use the following formula to choose the closest $K_\text{iter}$ points (indicate by $top\text{-}K$) and add them to the current group:
\begin{equation}
\mathcal{L}^t = \mathcal{L}^{t-1} \cup top\text{-}K_{p_{\hat{j}} \in P_i}\big(d(t,p_{\hat{j}}) \big)
\label{eq:inn_distance}
\end{equation}
This expansion continues until the number of points in $\mathcal{L}^t$ equals $K_\textbf{P}$.
As shown in Fig.\ref{fig:point_group} (c),
These newly incorporated points are all on the initial surface.
\end{itemize}
Fig.\ref{fig:point_group} (d) demonstrates (in \textcolor{blue}{blue}) that conventional KNN incorrectly merges points from two distinct surfaces, generating false positives.
In contrast, our IPG strategy preserves surface continuity in (e),
where previously problematic patches now exhibit normal feature characteristics,
effectively eliminating false detections within the marked bounding box region.

\subsection{Similar Neighborhood Aggregation with Multiple Degrees.}
\label{sec:3d_feature}
Given $F_\textbf{I}^{i, s}$ and $F_\textbf{P}^{i, s}$,
we propose SNAMD to capture anomalies of varying sizes,
we search the multi-scale neighborhoods for each patch.
Then, we design a similarity-weighted pooling method to aggregate neighborhood information, thereby preserving the discrimination of small anomalies.

In the 2D case, we reshape the vectorized patch tokens $F_\textbf{I}^{i, s}$ into $\sqrt{M_\textbf{I}} \times \sqrt{M_\textbf{I}} \times C_\textbf{I}$ grid to restore the spatial position information for the convenience of neighbor searching.
We extract $r \times r$ neighborhood features for a patch $m$ as $F_\textbf{I}^{i, s}(\mathcal{N}_r^m) \in \mathbb{R}^{|\mathcal{N}_r^m| \times C_\textbf{I}}$,
where $\mathcal{N}_r^m$ is the index list of $r \times r$ neighborhood patches of patch $m$, and $|\mathcal{N}_r^m|$ denotes its number.
Notably, a larger aggregation degree $r$ corresponds to a broader neighborhood, allowing the capture of larger anomaly regions.
In the 3D case, due to the irregularity of point clouds, 
for each 3D patch, we identify its $r$ nearest neighborhood by computing the Euclidean distances between the current patch center and all other patches in the 3D space.
The corresponding features of these neighboring patches are then defined as $F_\textbf{P}^{i, s}(\mathcal{N}_r^m) \in \mathbb{R}^{|\mathcal{N}_r^m| \times C_\textbf{P}}$.

\begin{figure}[!t]
\centering
\includegraphics[width=0.45\textwidth]{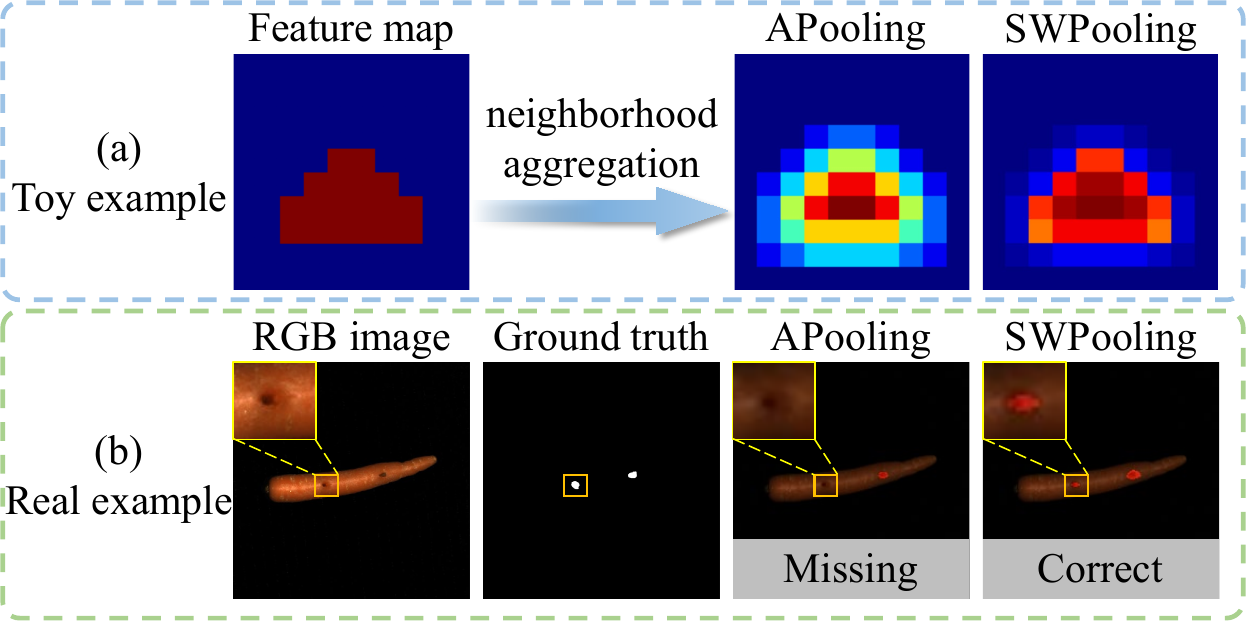}
\caption{
\textbf{Similarity-Weighted Pooling (SWPooling) Versus Average Pooling (APooling)}.
\emph{Top:}
One toy example represents feature maps aggregated by two aggregation methods,
where \textcolor[rgb]{0.0,0.0,0.5}{blue patches} and \textcolor[rgb]{0.5,0.0,0.0}{red patches} simulate normal and abnormal tokens, respectively.
\emph{Bottom:}
The visualization of segmentation results with SWPooling and APooling by one real example.}
\label{fig:agg_simi}
\end{figure}

\textbf{Similarity-Weighted Pooling}. Existing approaches \cite{CVPR2022patchcore,iclr2024musc} employ adaptive average pooling to aggregate the neighborhood features, which often dilutes small anomalies with surrounding normal patches by uniformly weighting neighbors,
particularly at larger $r$.
Therefore, we propose the similarity-weighted pooling (SWPooling),
which aggregates the most relevant neighborhood features to reduce the interference of irrelevant backgrounds.
Specifically, we calculate the similarity matrix of patch $m$ and all patches in the neighborhood $\mathcal{N}_r^m$, where higher similarity indicates lower interference.
The similarity matrix ${\Lambda}^{i, s}_{r, m} \in \mathbb{R}^{|\mathcal{N}_r^m| \times 1}$ is formulated as:

\begin{equation}
\label{image_simi}
{\Lambda}^{i, s}_{r, m}=\text{exp}({- \Vert F^{i, s}(\mathcal{N}_r^m)-F^{i, s}(m) \Vert_2})
\end{equation}
where $F^{i, s}(m) \in \mathbb{R}^{1 \times C}$ denotes the feature vector of patch $m$,
and $F^{i, s}(\mathcal{N}_r^m) \in \mathbb{R}^{|\mathcal{N}_r^m| \times C}$ represents the features of its $\mathcal{N}_r^m$ neighborhood.
The exponential function $\exp(\cdot)$ amplifies the importance of high-similarity patches.
Then, the similarity matrix performs the weighted average of the features within the neighborhood to generate the aggregated feature $\overline{F}^{i, s, r}(m)$,
\begin{equation}
\overline{F}^{i, s, r}(m)=\text{mean}({\Lambda}^{i, s}_{r, m} \odot F^{i, s}(\mathcal{N}_r^m))
\end{equation}
where $\odot$ represents element-wise multiplication.
Fig.\ref{fig:agg_simi} (a) compares standard average pooling (APooling) with our SWPooling.
APooling uniformly blends $|\mathcal{N}_r^m|$ neighborhoods and retains only $\frac{1}{|\mathcal{N}_r^m|}$ of the original patch's information.
This dilutes anomalies (\textcolor[rgb]{0.5,0.0,0.0}{red}) with surrounding normal features (\textcolor[rgb]{0.0,0.0,0.5}{blue}), causing missing detections in small anomalies.
In contrast, our SWPooling preserves local focus, suppressing these false negatives (yellow box in (b)).

For 2D patches,  we use multiple aggregation degrees, 
i.e.,  $r \in \{1, 3, 5\}$.
To optimize efficiency, we concatenate multi-scale features ($\overline{F}_\textbf{I}^{i, s, r}$ for $r \in \{1,3,5\}$) and compress them to $C_\textbf{I}$ dimensions,
yielding the final aggregated feature $\textbf{F}_\textbf{I}^{i, s} \in \mathbb{R}^{M_\textbf{I} \times C_\textbf{I}}$.
This optimization reduces subsequent mutual scoring operations, making MuSc-V2 ×5.6 faster than the conference version.
For 3D patches, to ensure surface-consistent aggregation, we restrict $r=1$ for high-curvature patches ($\mathcal{C} > \mathcal{C}_{thr}$), as in our IPG strategy.
Therefore, the neighborhood patches belong to the same surface.
We perform the same compression operation to obtain the aggregated 3D feature $\textbf{F}_\textbf{P}^{i, s}$.

\subsection{Multimodal Mutual Scoring}
\label{sec:m3sm}
According to the core observations mentioned above (normal patches across unlabelled samples could find many similar patches, while anomalies remain isolated),
we introduce a mutual scoring mechanism applicable to both 2D and 3D data.
This efficient mechanism generates high-quality patch-level anomaly scores through cross-sample comparison.

\begin{figure}[!t]
\centering
\includegraphics[width=0.49\textwidth]{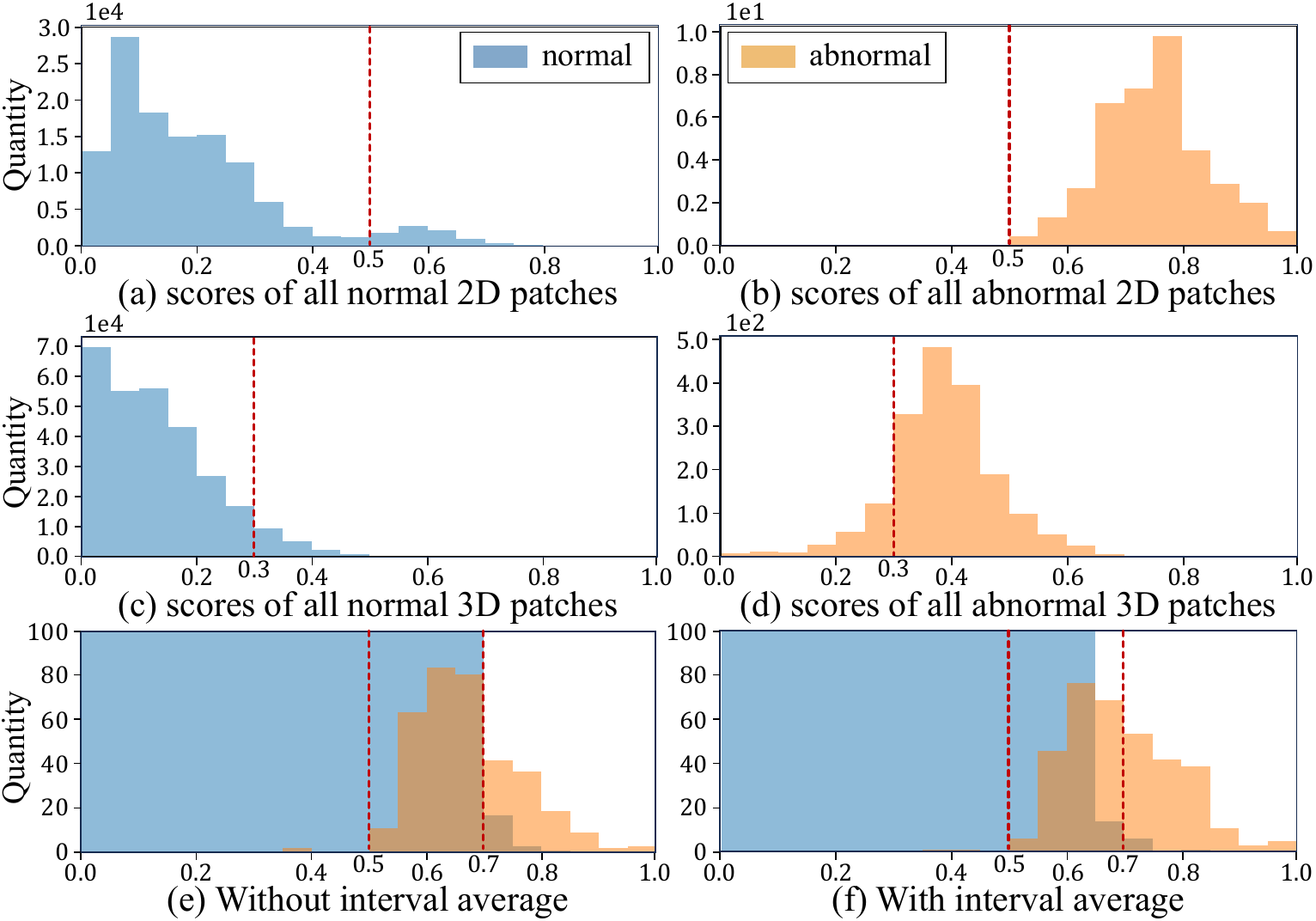}
\caption{
\textbf{(a-b)}
Score distributions $A_\textbf{I}^{i,s,m}$ for normal/abnormal 2D patches.
\textbf{(c-d)}
Corresponding score distributions for 3D patch.
\textbf{(e-f)}
Comparison of $\overline{a}_\textbf{I}^{i,s,m}$ distributions without/with Interval Average (IA) operation.
}
\label{fig:p2i}
\end{figure}

\textbf{Mutual Scoring Mechanism}.
Building upon the above discriminative aggregated features,
our MSM employs a novel paradigm where unlabeled samples mutually assign anomaly scores to each other.
Using 2D images as an illustrative case, we leverage each image in $\{\mathcal{D} \backslash I_i\}$ to assign a score for patch $m$ of image $I_i$ as follows,
\begin{equation}
\label{equ:rgg_msm}
a_\textbf{I}^{i,s,m}(I_j) = \min_{n} \Vert \textbf{F}_\textbf{I}^{i,s}(m)-\textbf{F}_\textbf{I}^{j,s}(n)\Vert_2
\end{equation}
where $(I_j)$ indicates that image $I_j \in \{\mathcal{D} \backslash I_i\}$ is employed for scoring.
If the patch token $\textbf{F}_\textbf{I}^{i,s}(m)$  of $I_i$ is similar to any patch token $\textbf{F}_\textbf{I}^{j,s}(n)$ of $I_j$, the image $I_j$ assigns a small anomaly score to $\textbf{F}_\textbf{I}^{i,s}(m)$.
In this way, each patch $m$ has a score set $A_\textbf{I}^{i,s,m}=\{a_\textbf{I}^{i,s,m}(I_j) | j \in [1, N], j \neq i \}$.
As shown in Fig.\ref{fig:p2i}, the histogram of $A_\textbf{I}^{i,s,m}$ for all normal (a) and abnormal (b) patches in $\mathcal{D}$ demonstrates the discriminative power.
These findings could generalize directly to 3D modal,
with Fig.\ref{fig:p2i} (c-d) demonstrating the same score distribution patterns for normal/abnormal 3D patches.

Our analysis above reveals that most unlabeled images in $\mathcal{D}$ assign lower scores to normal patches and higher scores to abnormal ones.
Therefore, simple average operation on all elements in the score list $A_\textbf{I}^{i,s,m}$ could effectively differentiate between normal and abnormal patches, shown in Fig. \ref{fig:p2i} (e).
However, minor overlaps in (e) occur when some normal patches exhibit appearance variations across images, resulting in elevated scores.
To mitigate this, we apply the Interval Average (IA) operation on the lowest $X\%$ of scores in $A_\textbf{I}^{i,s,m}$, suppressing outlier influences through:
\begin{equation}
\overline{a}_\textbf{I}^{i,s,m} = \frac{1}{K} \sum_{k \in [1,K]} a_\textbf{I}^{i,s,m}(\overline{I}_k)
\end{equation}
where $\overline{I}$ denotes images in the lowest $X\%$ score interval and $K$ is their count.
As shown in Fig.\ref{fig:p2i} (f), such a design reduces the normal/abnormal score overlap, particularly in [0.5, 0.7] compared to (e).
The final patch-level anomaly score $\textbf{a}_\textbf{I}^{i,m}$ combines multi-stage results as follows,
\begin{equation}
\textbf{a}_\textbf{I}^{i,m} = \frac{1}{S} \sum_{s \in [1,S]} \overline{a}_\textbf{I}^{i,s,m}
\end{equation}
yielding the patch-level anomaly score vector $\textbf{A}_\textbf{I}^{i} = [\textbf{a}_\textbf{I}^{i,1}, ..., \textbf{a}_\textbf{I}^{i,M_\textbf{I}}]^{\top}$.
Similarly, the patch-level anomaly score vector of the point cloud $P_i$ is denoted as $\textbf{A}_\textbf{P}^{i} = [\textbf{a}_\textbf{P}^{i,1}, ..., \textbf{a}_\textbf{P}^{i,M_\textbf{P}}]^{\top}$, where $\textbf{a}_\textbf{P}^{i,n}$ represents the anomaly score of the $n$-th 3D patch.

\begin{figure}[!t]
\centering
\includegraphics[width=0.49\textwidth]{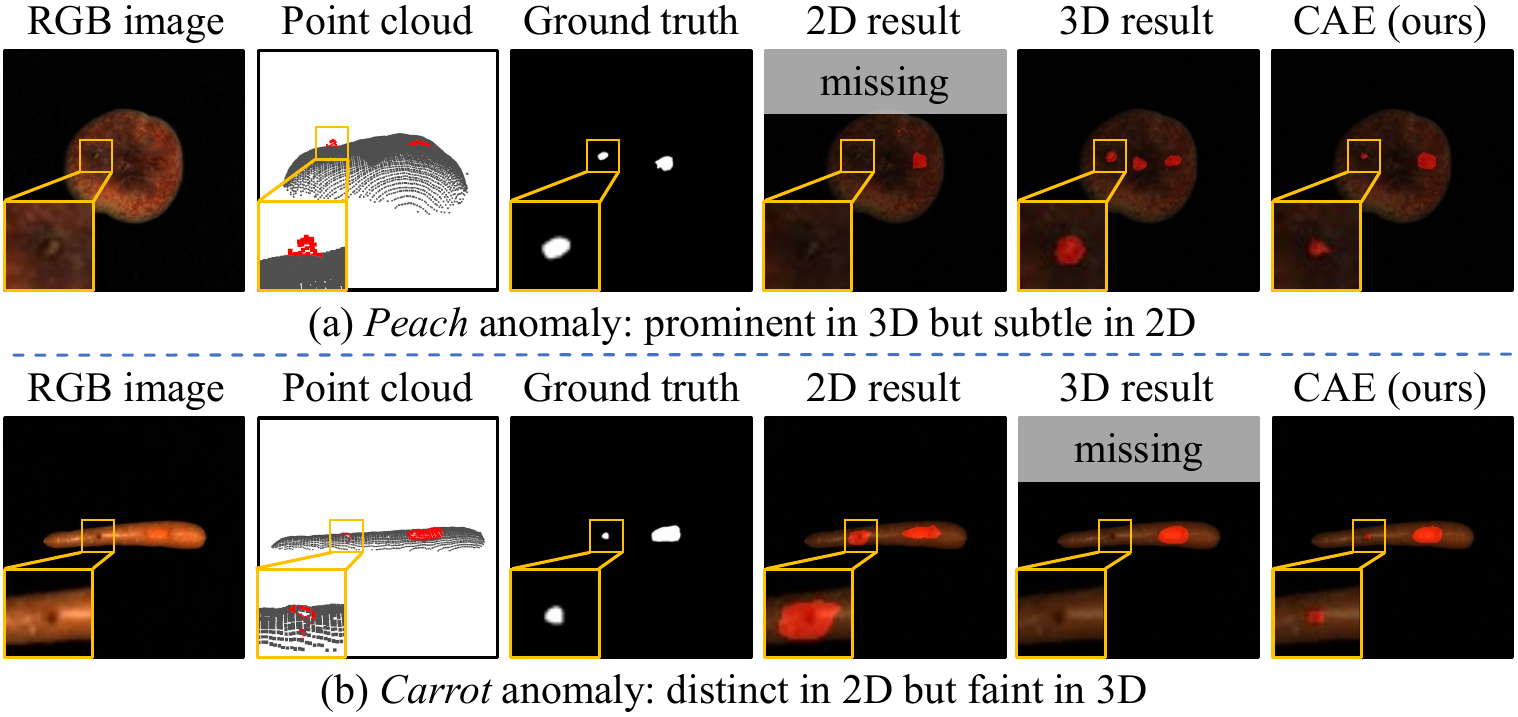}
\caption{
\textbf{Two examples whose anomalies exhibit single-modality prominence:}
(a) 3D-visible peach anomaly, (b) 2D-detectable carrot anomaly.
}
\label{fig:CMC}
\end{figure}

\textbf{Cross-modal Anomaly Enhancement}.
Our mutual scoring mechanism achieves strong patch-level anomaly detection within each modality,
yet faces limitations with modality-specific anomalies.
As Fig.\ref{fig:CMC} shows, \emph{peach} contamination (a) is prominent in 3D but subtle in 2D.
While the \emph{carrot} anomaly (b) is only significant in 2D.
These inherent data limitations constitute the theoretical bound for single-modal scoring.

To address this limitation, we propose the Cross-modal Anomaly Enhancement (CAE) module,
which augments anomalies invisible to single modal through cross-modal score fusion.
Using 2D data as an example, our approach begins with the mutual scoring mechanism (Eq.\ref{equ:rgg_msm}),
which computes patch scores $a_\textbf{I}^{i,s,m}(I_j)$ and $a_\textbf{P}^{i,s,n}(P_j)$ for each patch in 2D and 3D modals respectively.
The CAE then integrates these scores by two important steps.

\emph{Cross-modal alignment.}
To overcome the spatial misalignment of 2D and 3D patches,
we use the point coordinate and camera parameters to establish a projection list $\mathcal{P}_{i,m}$.
This mapping associates each 2D patch $m$ with its corresponding 3D points, thereby enabling accurate cross-modal mapping.
Then we average the 3D scores of all corresponding points to calculate the aligned 3D score $a_\textbf{P2I}^{i,s,n}(P_j)$ in 2D space as,
\begin{equation}
a_\textbf{P2I}^{i,s,m}(P_j) = \frac{1}{|\mathcal{P}_{i,m}|} \sum_{n \in \mathcal{P}_{i,m}} a_\textbf{P}^{i,s,n}(P_j)
\end{equation}
where $|\mathcal{P}_{i,m}|$ denotes the list length.
Scores with empty $\mathcal{P}_{i,m}$ (2D background patches) automatically are set to $0$.
To ensure the value range consistency of cross-modal scores, we rescale the projected 3D scores $A_\textbf{P2I}^{i,s,m}=\{a_\textbf{P2I}^{i,s,m}(P_k) | k \in [1, N], k \neq i \}$ into value range of 2D scores $A_\textbf{I}^{i,s,m}$.

\emph{Anomaly enhancement.}
To preserve anomalies detected in either modal, we perform $\max$ operation and fuse aligned 3D scores $a_\textbf{P2I}^{i,s,m}(P_j)$ with 2D scores $a_\textbf{I}^{i,s,m}(I_j)$ as,
\begin{equation}
a_\textbf{I}^{i,s,m}(I_j) \leftarrow a_\textbf{I}^{i,s,m}(I_j) + \lambda \max(a_\textbf{P2I}^{i,s,m}(P_j), a_\textbf{I}^{i,s,m}(I_j))
\end{equation}
where $\lambda \!=\! 1 \!-\! \text{std}(A_\textbf{P2I}^{i,s,m})$ is the confidence weight to measure each patch's reliability.
True anomalies exhibit low variance ($\lambda\!\rightarrow\!1$) due to consistent dissimilarity,
while false positives show higher variance from partial similarity with normal patches.
This design enhances anomalies and prevents cross-modal false positives, as validated in Fig.\ref{fig:CMC}.

We complete the remaining procedures of the mutual scoring mechanism to generate the patch-level anomaly score vectors $\textbf{A}_\textbf{I}^i$ for image $I_i$.
For 3D modal, we perform the same steps as in the 2D modal and obtain the patch-level anomaly score vectors $\textbf{A}_\textbf{P}^i$ for point cloud $P_i$.

\textbf{Multimodal Anomaly Segmentation.}
We convert patch-level anomaly scores to their original 2D/3D resolutions for segmentation evaluation.
For 2D data, the score vector $\textbf{A}_\textbf{I}^i \in \mathbb{R}^{M_\textbf{I} \times 1}$ is reshaped to $\sqrt{M_\textbf{I}} \times \sqrt{M_\textbf{I}} \times 1$ and upsampled.
For 3D data, we follow \cite{cvpr2023m3dm} to utilize inverse distance weight to interpolate scores to the point cloud.
This yields 2D anomaly segmentation result $\mathcal{A}_\textbf{I}^{i}$ and 3D anomaly segmentation result $\mathcal{A}_\textbf{P}^{i}$.
If both modals are available, the final segmentation combined their results through $\mathcal{A}^{i} = \mathcal{A}_\textbf{I}^{i} + \mathcal{A}_\textbf{P}^{i}$.

\textbf{Multimodal Anomaly Classification.}
Current AC/AS methods use the maximum score $c_i = \max(\mathcal{A}^{i})$ as the pixel-level anomaly classification (AC) score.
The AC score vector of samples in $\mathcal{D}$ is denoted as $\mathbf{C} = [c_1, ..., c_N]^{\top}$.
However $c_i$ is derived from the maximum value of the anomaly segmentation result,
it is sensitive to local noises and is easy to overlook weak anomalies.
In Sec. \ref{sec:rscon}, we propose the Re-Scoring with Constrained Neighborhood to mitigate these false classifications.

\begin{figure}[!t]
\centering
\includegraphics[width=0.49\textwidth]{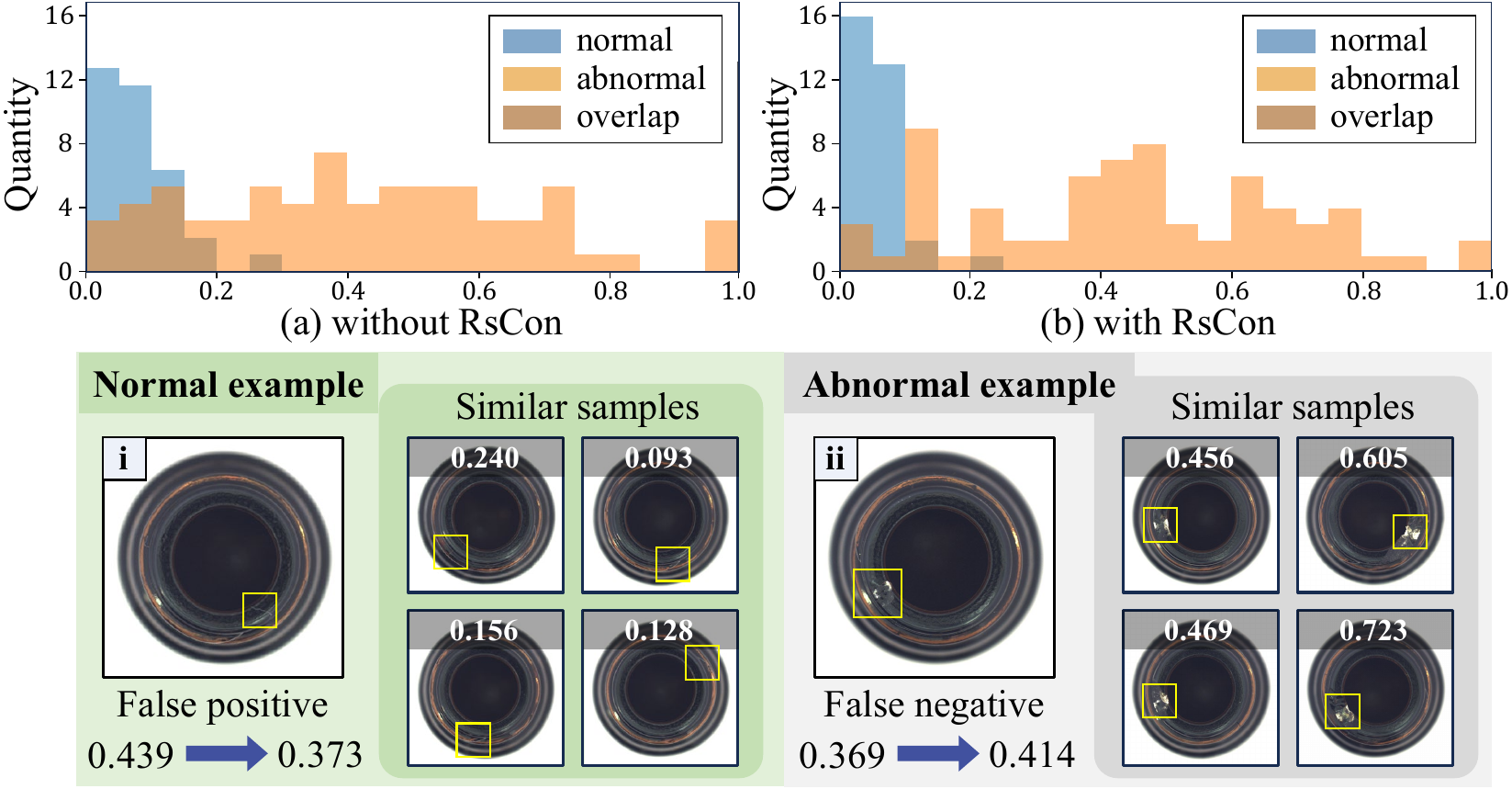}
\caption{
\textbf{Top:}
Histogram of anomaly classification scores of unlabeled samples before (a) and after (b) using our RsCon.
\textbf{Bottom:}
A normal example (i) and an abnormal example (ii) of RsCon.
}
\label{fig:rscon}
\end{figure}

\subsection{Re-Scoring with Constrained Neighborhood}
\label{sec:rscon}
While our mutual scoring mechanism effectively identifies most anomalies, it remains susceptible to both false positives and false negatives in certain challenging cases.
To mitigate these in anomaly classification, we introduce the concept of anomaly-salient feature,
which are extracted from the highest-scoring patch in the anomaly map.
We then calculate the similarity of these anomaly-salient features across different samples to calibrate scores $\mathbf{C}$.
As illustrated in Fig. \ref{fig:rscon}, a noisy but normal sample in (i) is incorrectly scored 0.439, while its similar samples have less noise and receive consistently lower scores (0.093–0.240).
The similar case holds in (ii), an abnormal sample with a subtle anomaly is assigned a lower score of 0.369, yet its similar samples have more visible anomalies and attain higher scores (0.456–0.723).
These observations reveal that score bias caused by local noise or weak anomalies could be corrected by referring to other similar samples.

According to the above observation and motivation,
we propose the re-scoring with constrained neighborhood (RsCon) to mitigate the above false classifications.
To calculate the anomaly-salient features, we extract them from the penultimate stage of the feature extractor as,
\begin{equation}
\mathcal{F}_\textbf{I}^i = \textbf{F}_\textbf{I}^{i,S-1}(\mathop{\arg\max}\limits_{m}(\textbf{a}_\textbf{I}^{i,m}))
\end{equation}
\begin{equation}
\mathcal{F}_\textbf{P}^i = \textbf{F}_\textbf{P}^{i,S-1}(\mathop{\arg\max}\limits_{m}(\textbf{a}_\textbf{P}^{i,m}))
\end{equation}
where $\mathcal{F}_\textbf{I}^i$ and $\mathcal{F}_\textbf{P}^i$ indicate 2D and 3D features respectively.
If both modals are available, we concatenate them into a multimodal feature $\mathcal{F}_i \in \mathbb{R}^{1 \times (C_\textbf{I}+C_\textbf{P})}$.
Then we construct an edge-weighted graph $\mathcal{G}=(\mathcal{V}, \mathcal{W})$ to build relationships in $\mathcal{D}$.
Each vertex $\mathcal{V}$ represents a sample, 
while the edge weights $\mathcal{W}$ are derived from a similarity matrix computed as ${\mathcal{W}_{i,j} = \mathcal{F}_{i} \cdot {\mathcal{F}_{j}}}$, where ${\cdot}$ means dot product.
With this sample-level similarity matrix, we employ manifold learning techniques \cite{ICCV2011sso, nips2003RDM, CVPR2012sd} to optimize the initial AC score $\textbf{C}$.

However, due to the close values in $\mathcal{W}_{i,\cdot}$, 
excessive features $\mathcal{F}_{j} (j \neq i)$ are propagated to $\mathcal{F}_{i}$, which make the AC accuracy decrease, as the experimental illustration in Sec.\ref{abl:rscon_module}.
Therefore, we design a Window Mask Operation (WMO) to constrain the number of samples.
The binary window mask matrix $\mathcal{M} \in \mathbb{R}^{N \times N}$ as follows, 
\begin{equation}
    \mathcal{M}(i,j) =
    \left \{ 
    \begin{array}{ll}
    1,~~\text{if}~~ O_{j} \in \mathcal{N}_{k}(O_i) \\[1mm]
    0,~~\text{otherwise},
    \end{array}\right.
\end{equation}
where $\mathcal{N}_{k}(O_i)$ indicates $k$ nearest samples of sample $O_i$.
Then the AC score $\textbf{C}$ is updated as, 
\begin{equation}
\hat{\textbf{C}} = \frac{1}{2}({D^{-1}}(\mathcal{M} \odot \mathcal{W})\textbf{C}+\textbf{C})
\label{eq:UpdatedACScore}
\end{equation}
where ${\hat{\textbf{C}} \in \mathbb{R}^{N \times 1}}$ is the optimized sample-level AC score vector,
and ${\odot}$ means element-wise multiplication.
$D$ normalizes $\mathcal{W}$ row-wise via ${D(i, i)}=\sum_{j=1}^{N}{\mathcal{M}{\odot}\mathcal{W}_{i,j}}$,
ensuring balanced contributions even from low-similarity neighbors.

\textbf{Discussions.}
To further explain the principle of the RsCon module clearly, we use it to optimize ${c_i}$ of the sample ${O_i}$ as an example.
According to the similarity matrix $\mathcal{W}$, we define the $k$-nearest neighbor to sample $O_i$ as $\{\hat{O}_1, ..., \hat{O}_k\}$ and their corresponding pixel-level AC scores as $\{\overline{c}_1, ..., \overline{c}_k\}$.
Then we use $\mathcal{M}\!\odot\!\mathcal{W}$ in Eq. \ref{eq:UpdatedACScore} to obtain their similarities to sample $O_i$ as $\{w_{i,1}, ..., w_{i,k}\}$.
The $D^{-1}$ normalizes these similarities and makes their sum equal to $1$.
The transformation results $\{\hat{w}^k_{i,1}, ..., \hat{w}^k_{i,k}\}$ are calculated as,
\begin{equation}
\begin{aligned}
&{\hat{w}^{k}_{i,j}} = \frac{w_{i,j}}{w_{i,1}+...+w_{i,k}},~\text{where}~{\hat{w}^{k}_{i,j}} \in \{\hat{w}^{k}_{i,1}, ..., \hat{w}^{k}_{i,k}\} \\
\end{aligned}
\end{equation}
Based on the above operations, using Eq. \ref{eq:UpdatedACScore} to optimize AC score ${c_i}$ of sample ${O_i}$ can be rewritten as,
\begin{equation}
\begin{aligned}
\hat{c}_i &= \frac{1}{2}((\hat{w}^k_{i,1}\overline{c}_1+...+\hat{w}^k_{i,k}\overline{c}_k) + c_i)\\
&=\frac{c_i}{2} + \frac{1}{2}\hat{w}^{k}_{i,1}\overline{c}_1+...+\frac{1}{2}\hat{w}^{k}_{i,k}\overline{c}_{k}\\
&=\frac{c_i}{2} + \frac{1}{2}\sum^{k}_{j=1}\hat{w}^k_{i,j}\overline{c}_{j}\\
\end{aligned}
\label{eq:app_cin}
\end{equation}
where $\hat{c}_i$ represents the optimized AC score of sample $O_i$.
Eq. \ref{eq:app_cin} shows that $\hat{c_i}$ is affected by anomaly classification scores in $k$-nearest neighbors.
The value of $c_i$ increases if sample $O_i$ has high-scoring $k$-neighbors (i.e., scores $\overline{c}_j \in \{\overline{c}_1, ..., \overline{c}_k\}$ are high), and vice versa.
Therefore sample $O_i$ with local noises could be corrected since its $k$ nearest neighbor samples have small AC scores.
As shown in Fig.\ref{fig:rscon} (b), the overlap between normal and abnormal AC scores reduces after applying RsCon compared with that in (a).

\begin{table*}
\vspace{-.4in}
\scriptsize
  \centering
  \caption{Quantitative comparisons on the \textbf{MVTec 3D-AD} and \textbf{Eyecandies} datasets. We compare our MuSc-V2 with some state-of-the-art zero-shot and few-shot methods. Bold indicates the best performance under zero-shot setting.
  The whole dataset is divided into $g$ subsets to simulate the small datasets.
  We report the mean and standard deviation over 10 random seeds, and the metrics decline is shown as $\downarrow$, which are marked in \colorbox{gray!20}{gray}.
  All metrics are in $\%$.}
  \setlength\tabcolsep{2.5pt}
    \begin{tabular}{@{}clllccccccc@{}}
    \toprule
    \multirow{2}{*}{Dataset} & \multirow{2}{*}{Method} & \multirow{2}{*}{Ref \& Year} & \multirow{2}{*}{Backbone} & \multicolumn{3}{c}{Anomaly Classification (AC)} & \multicolumn{4}{c}{Anomaly Segmentation (AS)} \\
    \cmidrule(l){5-7} \cmidrule(l){8-11} 
     &  &  &  & AUROC-cls & F1-max-cls & AP-cls & AUROC-seg & F1-max-seg & AP-seg & PRO@30\% \\
    \midrule[0.8pt]
    \multirow{10}{*}{MVTec 3D-AD \cite{VISAPP2022mvtec3d}} & ACR \cite{nips2023ACR} & NeurIPS'23 & WRN50 & 63.9 & 88.9 & 86.8 & 85.5 & 15.1 & 9.1 & 58.8 \\
    \multirow{10}{*}{(2D modal)} & APRIL-GAN \cite{arxiv2023APRILGAN} & CVPRW'23 & ViT-L-14-336 & 58.5 & 88.5 & 84.8 & 95.7 & 26.0 & 19.3 & 85.0 \\
    ~ & AnomalyCLIP \cite{ICLR2024anomalyclip} & ICLR'24 & ViT-L-14-336 & 65.1 & 88.7 & 87.9 & 96.2 & 33.5 & 28.1 & 83.6 \\
    ~ & AdaCLIP \cite{eccv2024adaclip} & ECCV'24 & ViT-L-14-336 & 74.8 & {89.9} & 91.8 & 97.6 & 41.7 & 36.5 & 61.5 \\
    ~ & VCP-CLIP \cite{eccv2024vcpclip} & ECCV'24 & ViT-L-14-336 & \textbf{76.3} & {89.9} & \textbf{92.6} & 97.7 & {42.5} & {38.3} & 92.2 \\
    ~ & FAPrompt \cite{iccv2025FAPrompt} & ICCV'25 & ViT-L-14-336 & 68.9 & 89.1 & 89.5 & 96.0 & 29.7 & 24.2 & 84.1 \\
    ~ & MuSc \cite{iclr2024musc} & Ours(ICLR'24) & ViT-L-14-336 & \textbf{76.3} & \textbf{90.3} & {92.3} & {97.9} & 41.0 & 36.4 & {93.1} \\
    ~ & MuSc-V2 & Ours & ViT-L-14-336 & {75.8} & \textbf{90.3(+0.4)} & 91.1 & \textbf{98.0(+0.1)} & \textbf{47.1(+4.6)} & \textbf{41.7(+3.4)} & \textbf{94.0(+0.9)} \\
    \cmidrule(l){2-11}
    ~ & MuSc\cite{iclr2024musc} & Ours(ICLR'24) & ViT-B-8 & 69.9 & 90.0 & 90.4 & 97.6 & 31.9 & 25.8 & 90.7 \\
    ~ & MuSc-V2 & Ours & ViT-B-8 & \textbf{75.7} & \textbf{90.5} & \textbf{92.6} & \textbf{98.2} & \textbf{40.4} & \textbf{34.7} & \textbf{93.1} \\
    \midrule
    \multirow{7}{*}{MVTec 3D-AD \cite{VISAPP2022mvtec3d}}
     & BTF (4-shot) \cite{cvpr2023btf} & CVPR'23 & FPFH & 66.3 & 88.5 & 88.4 & 96.8 & 32.6 & 28.7 & 88.3 \\
    \multirow{7}{*}{(3D modal)} & M3DM (4-shot) \cite{cvpr2023m3dm} & CVPR'23 & PT & 72.9 & 89.9 & 89.6 & 95.6 & 15.5 & 8.4 & 82.9 \\
    ~ & PointCLIPv2 \cite{iccv2023pointclipv2} & ICCV'23 & ViT-B-16 & 48.7 & 88.2 & 79.7 & 89.5 & 4.2 & 1.8 & 60.6 \\
    ~ & ULIP \cite{cvpr2023ulip} & CVPR'23 & PT & 59.2 & 88.5 & 83.9 & 89.8 & 4.7 & 2.2 & 61.2 \\
    ~ & ULIPv2 \cite{cvpr2024ulipv2} & CVPR'24 & PT & 63.5 & 89.1 & 86.3 & 91.0 & 4.8 & 2.3 & 64.9 \\
    ~ & PointAD \cite{nips2024pointad} & NeurIPS'24 & ViT-L-14-336 & {82.0} & {92.3} & {94.2} & {95.5} & {30.7} & {24.9} & {84.4} \\
    ~ & CMAD \cite{wacv20253Dzal} & CVPR'25 & ViT-H & 79.6 & - & 93.1 & - & - & - & - \\
    ~ & MuSc-V2 & Ours & PT & \textbf{83.7(+1.7)} & \textbf{92.5(+0.2)} & \textbf{94.4(+0.2)} & \textbf{97.1(+1.6)} & \textbf{45.9(+15.2)} & \textbf{44.4(+19.5)} & \textbf{88.4(+4.0)} \\
    \midrule
    \multirow{11}{*}{MVTec 3D-AD \cite{VISAPP2022mvtec3d}} & BTF (4-shot) \cite{cvpr2023btf} & CVPR'23 & WRN50+FPFH & 67.1 & 89.3 & 88.0 & 97.3 & 34.5 & 31.1 & 90.3 \\
    \multirow{11}{*}{(Multimodal)} & M3DM (4-shot) \cite{cvpr2023m3dm} & CVPR'23 & ViT-B-8+PT & 78.5 & 91.0 & 91.9 & 97.9 & 35.1 & 32.1 & 92.4 \\
    ~ & CFM (4-shot) \cite{cvpr2024cfm} & CVPR'24 & ViT-B-8+PT & 77.3 & 91.2 & 92.4 & 98.3 & 42.4 & 40.5 & 94.0 \\
    ~ & PointCLIPv2 \cite{iccv2023pointclipv2} & ICCV'23 & ViT-B-16 & 66.4 & 89.2 & 88.6 & 95.4 & 21.8 & 13.2 & 79.1 \\
    ~ & ULIP \cite{cvpr2023ulip} & CVPR'23 & ViT-B-16+PT & 61.5 & 89.2 & 85.8 & 95.0 & 21.1 & 12.6 & 78.0 \\
    ~ & ULIPv2 \cite{cvpr2024ulipv2} & CVPR'24 & ViT-B-16+PT & 59.8 & 89.0 & 84.8 & 95.1 & 21.0 & 12.3 & 78.3 \\
    ~ & PointAD \cite{nips2024pointad} & NeurIPS'24 & ViT-L-14-336 & 86.9 & 92.2 & 96.1 & 97.2 & 37.2 & 31.0 & 90.2 \\
    ~ & 3DzAL \cite{wacv20253Dzal} & WACV'25 & WRN50+PointNet++ & 64.9 & - & - & - & - & - & 84.8 \\
    ~ & MuSc-V2 & Ours & ViT-B-8+PT & \textbf{88.1(+1.2)} & \textbf{93.0(+0.8)} & \textbf{96.8(+0.7)} & \textbf{99.0(+1.8)} & \textbf{54.6(+17.4)} & \textbf{54.7(+23.7)} & \textbf{97.0(+6.8)} \\
    
    ~ & \cellcolor{gray!20}MuSc-V2 ($g$=2) & \cellcolor{gray!20}Ours & \cellcolor{gray!20}ViT-B-8+PT & \cellcolor{gray!20}87.6±0.2\textbf{\tiny $\downarrow$0.5} & \cellcolor{gray!20}92.7±0.2\textbf{\tiny $\downarrow$0.3} & \cellcolor{gray!20}96.6±0.1\textbf{\tiny $\downarrow$0.2} & \cellcolor{gray!20}99.0±0.0\textbf{\tiny $\downarrow$0.0} & \cellcolor{gray!20}54.3±0.2\textbf{\tiny $\downarrow$0.3} & \cellcolor{gray!20}54.4±0.2\textbf{\tiny $\downarrow$0.3} & \cellcolor{gray!20}96.9±0.0\textbf{\tiny $\downarrow$0.1} \\
    
    ~ & \cellcolor{gray!20}MuSc-V2 ($g$=3) & \cellcolor{gray!20}Ours & \cellcolor{gray!20}ViT-B-8+PT & \cellcolor{gray!20}87.2±0.3\textbf{\tiny $\downarrow$0.9} & \cellcolor{gray!20}92.6±0.2\textbf{\tiny $\downarrow$0.4} & \cellcolor{gray!20}96.4±0.1\textbf{\tiny $\downarrow$0.4} & \cellcolor{gray!20}99.0±0.0\textbf{\tiny $\downarrow$0.0} & \cellcolor{gray!20}53.6±0.2\textbf{\tiny $\downarrow$1.0} & \cellcolor{gray!20}53.7±0.2\textbf{\tiny $\downarrow$1.0} & \cellcolor{gray!20}96.8±0.1\textbf{\tiny $\downarrow$0.2} \\
    ~ & MuSc-V2 & Ours & ViT-L-14-336+PT & 89.9 & 93.6 & 97.2 & 98.8 & 58.8 & 60.0 & 96.6 \\
    \midrule[0.8pt]
    \multirow{9}{*}{Eyecandies \cite{accv2022eyecandies}} & APRIL-GAN \cite{arxiv2023APRILGAN} & CVPRW'23 & ViT-L-14-336 & 62.2 & 68.7 & 65.4 & 93.4 & 23.7 & 17.6 & 77.0 \\
    \multirow{9}{*}{(2D modal)} & AnomalyCLIP \cite{ICLR2024anomalyclip} & ICLR'24 & ViT-L-14-336 & 73.8 & 74.4 & 75.7 & 91.1 & 26.3 & 17.7 & 73.7 \\
    ~ & AdaCLIP \cite{eccv2024adaclip} & ECCV'24 & ViT-L-14-336 & 74.3 & 73.5 & 75.2 & 96.9 & 34.6 & 28.6 & 42.0 \\
    ~ & VCP-CLIP \cite{eccv2024vcpclip} & ECCV'24 & ViT-L-14-336 & 73.7 & 74.6 & 75.5 & {97.1} & 35.4 & 30.1 & 87.2 \\
    ~ & FAPrompt \cite{iccv2025FAPrompt} & ICCV'25 & ViT-L-14-336 & 74.4 & 74.6 & 75.6 & 93.8 & 26.3 & 20.1 & 76.7 \\
    ~ & MuSc \cite{iclr2024musc} & Ours(ICLR'24) & ViT-L-14-336 & {78.0} & {77.5} & {80.4} & \textbf{97.3} & {37.9} & {33.8} & {88.9} \\
    ~ & MuSc-V2 & Ours & ViT-L-14-336 & \textbf{85.1(+7.1)} & \textbf{84.8(+7.3)} & \textbf{87.4(+7.0)} & 96.9 & \textbf{44.0(+6.1)} & \textbf{38.9(+5.1)} & \textbf{89.4(+0.5)} \\
    \cmidrule(l){2-11}
     ~ & MuSc \cite{iclr2024musc} & Ours(ICLR'24) & ViT-B-8 & 73.4 & 74.2 & 73.8 & 96.6 & 30.3 & 24.2 & 84.9 \\
    ~ & MuSc-V2 & Ours & ViT-B-8 & \textbf{85.1} & \textbf{81.8} & \textbf{86.6} & \textbf{97.4} & \textbf{42.0} & \textbf{36.8} & \textbf{89.0} \\
    \midrule
    \multirow{3}{*}{Eyecandies \cite{accv2022eyecandies}} & BTF (4-shot) \cite{cvpr2023btf} & CVPR'23 & FPFH & 64.2 & 69.9 & 67.5 & 85.8 & 29.0 & 21.8 & 62.3 \\
    \multirow{3}{*}{(3D modal)} & M3DM (4-shot) \cite{cvpr2023m3dm} & CVPR'23 & PT & 64.9 & 69.8 & 66.6 & 88.3 & 9.0 & 6.0 & 63.1 \\
    ~ & PointAD \cite{nips2024pointad} & NeurIPS'24 & ViT-L-14-336 & \textbf{69.1} & 71.2 & \textbf{73.8} & \textbf{92.1} & 24.9 & 15.9 & \textbf{71.3} \\
    ~ & MuSc-V2 & Ours & PT & \textbf{69.1} & \textbf{72.1(+0.9)} & 71.7 & 89.8 & \textbf{28.1(+3.2)} & \textbf{21.4(+5.5)} & 66.9 \\
    \midrule
    \multirow{7}{*}{Eyecandies \cite{accv2022eyecandies}} & BTF (4-shot) \cite{cvpr2023btf} & CVPR'23 & WRN50+FPFH & 65.3 & 68.8 & 67.0 & 92.7 & 24.2 & 19.3 & 74.3 \\
    \multirow{7}{*}{(Multimodal)} & M3DM (4-shot) \cite{cvpr2023m3dm} & CVPR'23 & ViT-B-8+PT & 73.5 & 75.3 & 75.7 & 96.1 & 32.4 & 27.5 & 82.8 \\
    ~ & CFM (4-shot) \cite{cvpr2024cfm} & CVPR'24 & ViT-B-8+PT & 71.6 & 72.8 & 73.9 & 96.4 & 32.9 & 29.0 & 82.5 \\
    ~ & PointAD \cite{nips2024pointad} & NeurIPS'24 & ViT-L-14-336 & 77.7 & 76.4 & 80.4 & 95.3 & 30.5 & 22.5 & 84.3 \\
    ~ & MuSc-V2 & Ours & ViT-B-8+PT & \textbf{83.9(+6.2)} & \textbf{82.9(+6.5)} & \textbf{86.1(+5.7)} & \textbf{97.5(+2.2)} & \textbf{44.7(+14.2)} & \textbf{41.8(+19.3)} & \textbf{90.1(+5.8)} \\
    ~ & \cellcolor{gray!20}MuSc-V2 ($g$=2) & \cellcolor{gray!20}Ours & \cellcolor{gray!20}ViT-B-8+PT & \cellcolor{gray!20}83.5±0.6\textbf{\tiny $\downarrow$0.4} & \cellcolor{gray!20}82.6±0.4\textbf{\tiny $\downarrow$0.3} & \cellcolor{gray!20}85.8±0.4\textbf{\tiny $\downarrow$0.3} & \cellcolor{gray!20}97.4±0.0\textbf{\tiny $\downarrow$0.1} & \cellcolor{gray!20}44.4±0.2\textbf{\tiny $\downarrow$0.3} & \cellcolor{gray!20}41.3±0.2\textbf{\tiny $\downarrow$0.5} & \cellcolor{gray!20}89.8±0.1\textbf{\tiny $\downarrow$0.2} \\
    ~ & \cellcolor{gray!20}MuSc-V2 ($g$=3) & \cellcolor{gray!20}Ours & \cellcolor{gray!20}ViT-B-8+PT & \cellcolor{gray!20}83.4±0.7\textbf{\tiny $\downarrow$0.5} & \cellcolor{gray!20}82.2±0.7\textbf{\tiny $\downarrow$0.7} & \cellcolor{gray!20}85.5±0.5\textbf{\tiny $\downarrow$0.6} & \cellcolor{gray!20}97.4±0.1\textbf{\tiny $\downarrow$0.1} & \cellcolor{gray!20}43.8±0.3\textbf{\tiny $\downarrow$0.9} & \cellcolor{gray!20}40.6±0.3\textbf{\tiny $\downarrow$1.2} & \cellcolor{gray!20}89.8±0.3\textbf{\tiny $\downarrow$0.3} \\
    ~ & MuSc-V2 & Ours & ViT-L-14-336+PT & 85.9 & 84.4 & 87.6 & 97.3 & 46.2 & 42.6 & 90.9 \\
    \bottomrule
  \end{tabular}
  \label{tab:comp_datasets_all}
\end{table*}

\section{Experiments}
\subsection{Experimental setting}
\subsubsection{Datasets}
We conduct experiments on two multimodal industrial datasets (MVTec 3D-AD \cite{VISAPP2022mvtec3d} and Eyecandies \cite{accv2022eyecandies}) and four 2D-only datasets (MVTec AD \cite{CVPR2019mvtec}, VisA \cite{ECCV2022VisA}, MPDD \cite{icumt2021mpdd}, and BTAD \cite{ISIE2021VT-ADL}).
MVTec 3D-AD consists of 10 product categories with 41 types of anomalies,
where 3D points are stored as XYZ maps matching the RGB image resolution.
Eyecandies provides synthetic data for 10 categories of candies, cookies and sweets,
including depth maps and camera parameters for 3D reconstruction.
For 2D-only datasets, MVTec AD contains 15 object/texture categories, VisA has 12 objects across 3 domains, the MPDD and BTAD datasets each contain 6 and 3 products respectively.
All datasets provide normal and abnormal samples in the unlabeled test set.
Notably, our method could operate effectively without requiring the full dataset.
To verify its stability and robustness of on smaller datasets, we randomly partition the whole unlabeled dataset into $g$ subsets, each containing an equal number of samples, thereby simulating the behavior on small‑scale datasets.
During evaluation, we collect the anomaly maps from all subsets and compute the metrics jointly to ensure a fair comparison.

\subsubsection{Evaluation Metrics}
For image-level anomaly classification, we report 3 widely used metrics:
the Area Under Receiver Operator Characteristic curve (AUROC), Average Precision (AP), and F1-score at optimal threshold (F1-max).
For pixel-level anomaly segmentation, we evaluate with pixel-wise AUROC, F1-max, AP, and Per-Region Overlap with 30\% FPR (PRO@30\%) \cite{CVPR2019mvtec}.
All metrics above are calculated using official implementations.

\subsubsection{Implementation Details}
Following current multimodal anomaly detection methods \cite{cvpr2023m3dm,cvpr2023btf,cvpr2024cfm}, we use DINO ViT-B-8 \cite{iccv2021dino} for 2D feature extraction and Point Transformer \cite{ICCV2021pointTransformer} (pre-trained with Point-MAE \cite{eccv2022pointmae}) for 3D feature extraction.
For fair comparisons with some 2D-only methods, we also include ViT-L-14-336 pretrained with CLIP \cite{ICML2021CLIP}.
Both ViT and PT architectures are divided into 3 stages ($S$=3).
The input images are resized to 224×224, while point clouds are clustered into 1024 groups of 128 points each.
\textit{Note that, to ensure robust evaluation of subset partitioning,
we conduct experiments across 10 random seeds and report averaged results.}
Key hyperparameters in our MuSc-V2 include:
IPG's iterative point increment $K_\text{iter}$=80 and curvature threshold $\mathcal{C}_{thr}$=0.01,
SNAMD's aggregation degrees $r \!\in\! \{1,3,5\}$,
MSM's minimum 30\% score interval for IA,
and RsCon's window size $k$=7.
All hyperparameters above are consistent across all datasets.

\subsubsection{Competing methods}
For 2D modal, we compare with some state-of-the-art zero-shot approaches, e.g. APRIL-GAN \cite{arxiv2023APRILGAN}, AnomalyCLIP \cite{ICLR2024anomalyclip}, ACR \cite{nips2023ACR}, AdaCLIP \cite{eccv2024adaclip}, VCP-CLIP \cite{eccv2024vcpclip} and RareCLIP \cite{iccv2025RareCLIP}.
For the CLIP-based methods, we load official checkpoints for inference.
In multimodal scenarios, we evaluate against PointCLIPv2 \cite{iccv2023pointclipv2}, ULIP \cite{cvpr2023ulip}, ULIPv2 \cite{cvpr2024ulipv2}, PointAD \cite{nips2024pointad}, 3DzAL \cite{wacv20253Dzal} and CMAD \cite{cvpr2025CMAD}.
Since 3DzAL and CMAD have not been open-sourced, we only report the results from their official papers, with unavailable results denoted by `-'.
In addition, we also compare few-shot methods, such as M3DM \cite{cvpr2023m3dm}, CFM \cite{cvpr2024cfm} and BTF \cite{cvpr2023btf}.

\begin{table*}
\vspace{-.4in}
\scriptsize
  \centering
  \caption{Quantitative comparisons on four 2D-only datasets: \textbf{MVTec AD}, \textbf{VisA}, \textbf{MPDD}, and \textbf{BTAD}. Bold indicates the best performance.}
  \setlength\tabcolsep{2.5pt}
    \begin{tabular}{@{}lcccccc|lcccccc@{}}
    \toprule
    \multirow{2}{*}{Method} & \multicolumn{3}{c}{Anomaly Classification} & \multicolumn{3}{c|}{Anomaly Segmentation} & \multirow{2}{*}{Method} & \multicolumn{3}{c}{Anomaly Classification} & \multicolumn{3}{c}{Anomaly Segmentation} \\
    \cmidrule(l){2-4} \cmidrule(l){5-7} \cmidrule(l){9-11} \cmidrule(l){12-14}
    ~ & AUROC-cls & F1-max-cls & AP-cls & AUROC-seg & F1-max-seg & AP-seg & ~ & AUROC-cls & F1-max-cls & AP-cls & AUROC-seg & F1-max-seg & AP-seg \\
    \midrule[0.8pt]
    \multicolumn{7}{c|}{MVTec AD \cite{CVPR2019mvtec}} & \multicolumn{7}{c}{MPDD \cite{icumt2021mpdd}} \\
    \midrule[0.2pt]
    AnomalyCLIP \cite{ICLR2024anomalyclip} & 91.5 & 92.8 & 96.2 & 91.1 & 39.1 & 34.5 & AnomalyCLIP & 77.0 & 79.9 & 82.0 & 96.5 & 34.2 & 28.9 \\
    AdaCLIP \cite{eccv2024adaclip} & 89.2 & 90.6 & 95.7 & 88.7 & 43.4 & 41.6 & AdaCLIP & 76.0 & 82.5 & 78.9 & 96.1 & 34.9 & 31.9 \\
    VCP-CLIP \cite{eccv2024vcpclip} & 92.1 & 91.7 & 95.5 & 92.0 & 49.4 & 49.1 & VCP-CLIP & 69.3 & 78.1 & 72.9 & 96.8 & 31.2 & 27.8 \\
    RareCLIP \cite{iccv2025RareCLIP} & 91.5 & 92.9 & 96.6 & 91.5 & 47.5 & 46.1 & RareCLIP & 78.0 & 82.7 & \textbf{83.3} & 95.8 & 34.8 & 31.5 \\
    MuSc \cite{iclr2024musc} & \textbf{97.8} & {97.5} & \textbf{99.1} & \textbf{97.3} & {62.6} & {62.7} & MuSc & 78.2 & \textbf{83.8} & 79.3 & \textbf{97.5} & 36.3 & \textbf{33.0} \\
    MuSc-V2 (Ours) & 97.7 & \textbf{97.6} & \textbf{99.1} & \textbf{97.3} & \textbf{63.1} & \textbf{63.5} & MuSc-V2 (Ours) & \textbf{79.0} & 83.6 & 78.7 & \textbf{97.5} & \textbf{37.7} & 32.6 \\
    \midrule[0.8pt]
    \multicolumn{7}{c|}{VisA \cite{ECCV2022VisA}} & \multicolumn{7}{c}{BTAD \cite{ISIE2021VT-ADL}} \\
    \midrule[0.2pt]
    AnomalyCLIP & 82.1 & 80.7 & 85.4 & 95.5 & 28.3 & 21.3 & AnomalyCLIP & 88.3 & 83.8 & 87.3 & 94.2 & 49.7 & 45.5 \\
    AdaCLIP & 85.8 & 83.1 & 87.6 & 95.5 & 37.7 & 31.1 & AdaCLIP & 88.6 & 88.2 & 93.3 & 92.1 & 51.7 & 47.8 \\
    VCP-CLIP & 83.8 & 81.5 & 87.3 & 95.7 & 34.7 & 30.1 & VCP-CLIP & 89.3 & 88.8 & 92.6 & 93.8 & 48.5 & 45.1 \\
    RareCLIP & 86.1 & 83.1 & 89.0 & 95.7 & 33.5 & 27.0 & RareCLIP & 91.6 & 89.1 & 93.4 & 91.7 & 54.1 & 50.1 \\
    MuSc & \textbf{92.8} & \textbf{89.5} & \textbf{93.5} & \textbf{98.8} & {48.8} & {45.1} & MuSc & 94.8 & 94.6 & 97.5 & 97.3 & 57.5 & 54.1 \\
    MuSc-V2 (Ours) & 91.9 & {88.5} & {92.8} & \textbf{98.8} & \textbf{50.4} & \textbf{46.1} & MuSc-V2 (Ours) & \textbf{96.5} & \textbf{96.1} & \textbf{99.1} & \textbf{97.5} & \textbf{59.0} & \textbf{57.3} \\
    \bottomrule
  \end{tabular}
  \label{tab:comp_datasets_rgb}
\end{table*}

\begin{figure}[!t]
\centering
\includegraphics[width=0.47\textwidth]{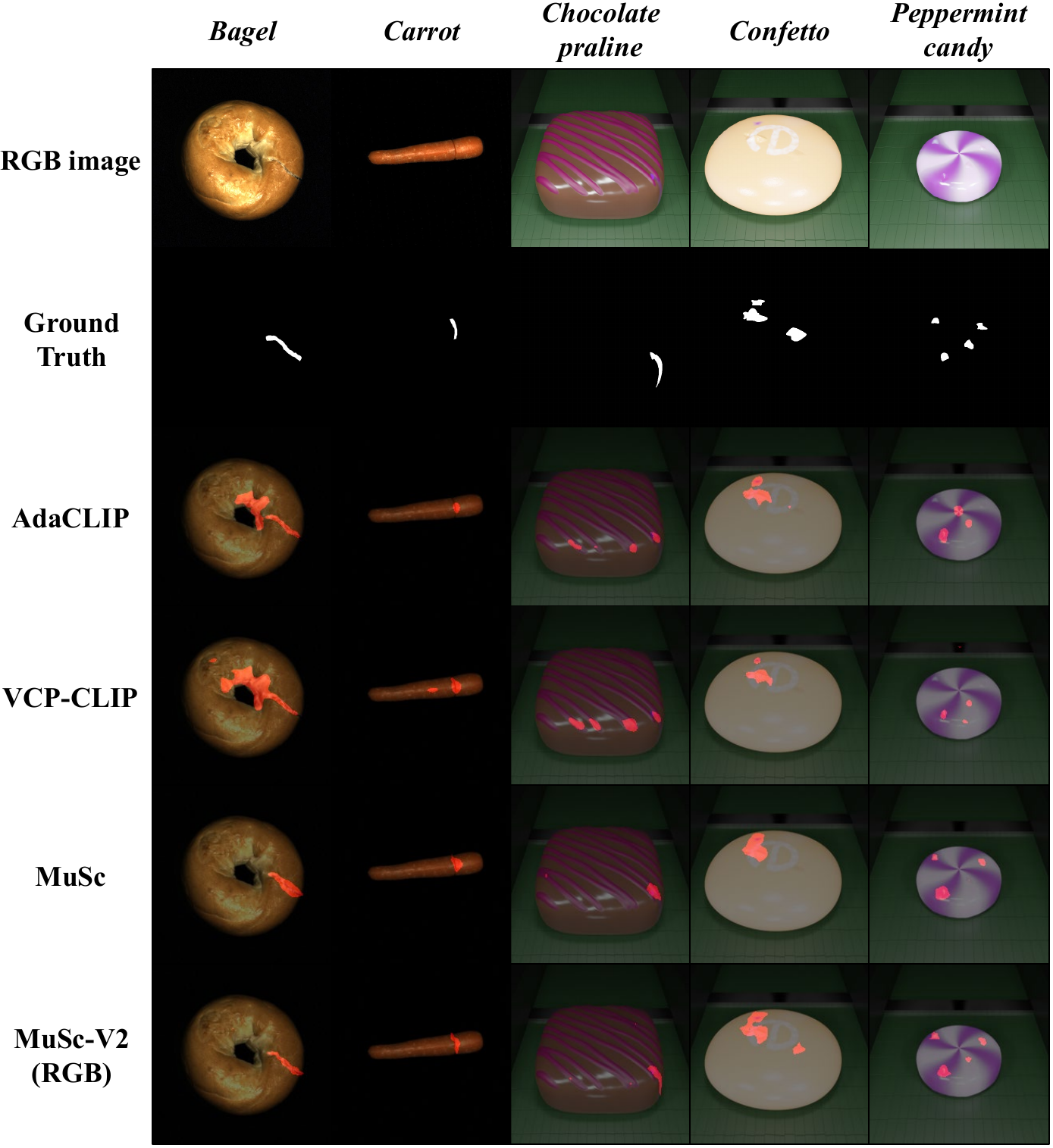}
\caption{
Visualization of anomaly segmentation on MVTec 3D-AD and Eyecandies under 2D modal.
All methods use ViT-L-14-336 extract features.
}
\label{fig:vis_rgb}
\end{figure}

\subsection{Quantitative results}
In Table \ref{tab:comp_datasets_all}, we compare our MuSc-V2 with state-of-the-art zero-shot and few-shot methods on MVTec 3D-AD and Eyecandies datasets.
We report the anomaly classification and segmentation results across 2D, 3D and multimodal settings.
MuSc-V2 achieves superior performance in most metrics for all modals.
Notably, it outperforms the second-best zero-shot method PointAD\cite{nips2024pointad} by $\textbf{23.7\%}$ and $\textbf{19.3\%}$ AP for anomaly segmentation on these datasets.
For anomaly classification, MuSc-V2 achieves $1.2\%$ and $\textbf{6.2\%}$ AUROC gains on both datasets.
When partitioning the original dataset into $g \in \{2,3\}$ subsets, both AC and AS metrics show minimal degradation (\colorbox{gray!20}{$\downarrow$}): at most 1.0\% on MVTec 3D-AD and 1.2\% on Eyecandies.
The slightly larger impact on Eyecandies stems from its limited sample size (50 samples per product).
We report these performances as mean±std, with maximum standard deviations of 0.3 on MVTec 3D-AD and 0.7 on Eyecandies.
These small variations confirm the insensitivity to different partitioning schemes and adaptability to diverse dataset compositions.
When evaluated against 2D-only zero-shot methods \cite{ICLR2024anomalyclip,eccv2024adaclip,eccv2024vcpclip,iccv2025RareCLIP} on four 2D-only datasets (Table \ref{tab:comp_datasets_rgb}),
MuSc-V2 demonstrates significant improvements, even surpassing its previous version MuSc \cite{iclr2024musc} across most metrics.
However, some background regions on the VisA dataset contain subtle noise.
Although these are not true anomalies, they are captured by the SNAMD module due to its sensitivity to subtle anomalies.
This affects the AC metric, while the AS metric remains nearly unchanged.
The reason is that the false positives occur only in very small regions, resulting in a few pixels with high anomaly scores.
This causes image‑level over‑detection but has only a negligible impact on pixel‑level evaluation.
In addition, products in the MPDD dataset exhibit more pose variations, making it harder to find similar normal regions across images with different poses.
While some existing methods achieve a slightly higher AP-cls, our method still delivers competitive or superior results on all other metrics.

\begin{figure*}[t]
\vspace{-.1in}
\begin{center}
\includegraphics[width=0.96\textwidth]{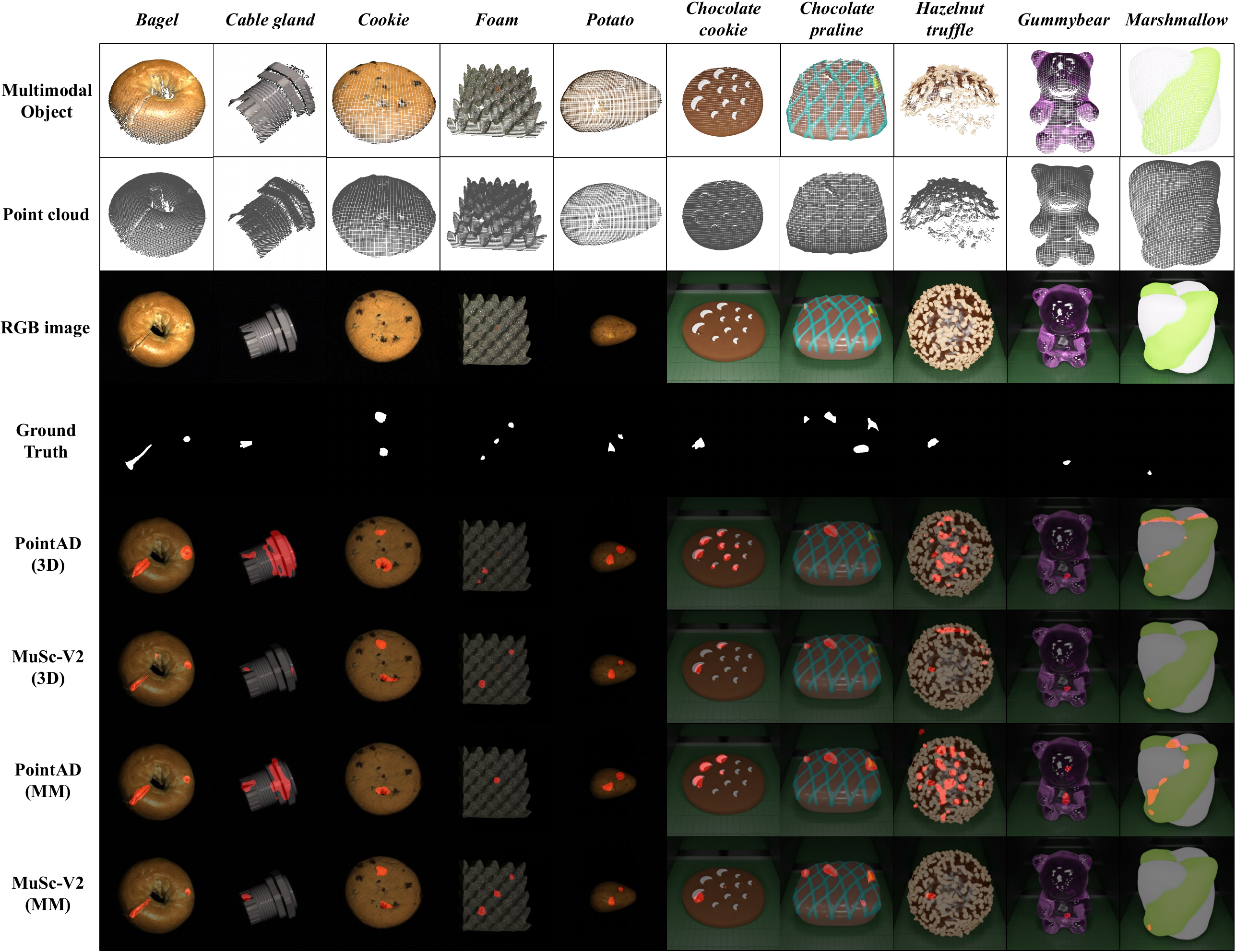}
\vspace{-1mm} 
\caption{
Visualization of anomaly segmentation results on MVTec 3D-AD and Eyecandies benchmarks.
3D modal and multimodal (MM) results are displayed.
}
\label{fig:vis_3d}
\end{center}
\vspace{-10pt}
\end{figure*}

\subsection{Qualitative results}
We visualize the multimodal anomaly segmentation results in Fig. \ref{fig:vis_3d}.
Compared with other zero-shot methods, MuSc-V2 generates fewer false positives, e.g., \textit{cable gland} and \textit{marshmallow}.
Our method also avoids false negatives common in multi-view rendering approaches like PointAD \cite{nips2024pointad},
particularly for objects with angles of view occlusion (\textit{foam}).
By detecting in 3D point cloud directly, we achieve more precise segmentation in complex cases (\textit{bagel}, \textit{potato} and \textit{gummybear}).
The 2D results in Fig. \ref{fig:vis_rgb} demonstrate that MuSc-V2 reduces false positives (\textit{chocolate praline}) and false negatives (\textit{confetto}, \textit{peppermint candy}),
and yields finer results than its previous version in \textit{carrot} and \textit{chocolate praline}.

\subsection{Ablation study}

\begin{table}[!t]
    \centering
    \caption{The ablation of the IPG strategy.
    We report the anomaly classification and segmentation results.
    All metrics are in $\%$.}
  \label{tab:INN_module}
  \setlength{\tabcolsep}{1.2mm}{
    \begin{tabular}{c|c|cccc}
        \toprule
        Dataset & Setting & F1-max-cls & AP-cls & F1-max-seg & AP-seg \\ 
        \midrule 
        \multirow{2}{*}{MVTec 3D-AD} & w/o IPG  & 92.5 & 96.7 & 54.6 & 54.6 \\ 
        ~ & w IPG & \textbf{93.0} & \textbf{96.8} & 54.6 & \textbf{54.7} \\
        \midrule
        \multirow{2}{*}{Eyecandies} & w/o IPG  & 82.5 & 85.7 & 44.7 & 41.8 \\ 
        ~ & w IPG & \textbf{82.9} & \textbf{86.1} & 44.7 & 41.8 \\
        \bottomrule
    \end{tabular}
    }
\end{table}

\subsubsection{Effectiveness of the IPG strategy}
\label{abl:INN_module}
In Table \ref{tab:INN_module}, we conduct experiments to validate the effectiveness of our IPG strategy.
This brings 0.5$\%$ and 0.4$\%$ F1-max-cls gains on MVTec 3D-AD and Eyecandies datasets respectively.
Since the ``groups containing discontinuous surfaces" typically occupy small local regions,
their improvements on anomaly segmentation metrics are small (about 0.1$\%$).
However, it is effective for sample-level anomaly classification by reducing false positives in normal point clouds.

\begin{table}[!t]
    \centering
    \caption{The ablation of four important modules in our multimodal MSM.
    We report the AC and AS results.
    All metrics are in $\%$.}
  \label{tab:m3sm_module}
  \setlength{\tabcolsep}{1.2mm}{
    \begin{tabular}{c|l|cccc}
        \toprule
        Dataset & Setting & F1-max-cls & AP-cls & F1-max-seg & AP-seg \\ 
        \midrule 
        \multirow{4}{*}{MVTec 3D-AD} & w/o IA  & 92.4 & 96.1 & 52.5 & 52.5 \\ 
        ~ & w/o CAE  & 91.7 & 96.0 & 52.5 & 52.1 \\
        ~ & w/o $\lambda$ & 92.4 & 96.7 & 54.4 & 54.3 \\
        ~ & Ours & \textbf{93.0} & \textbf{96.8} & \textbf{54.6} & \textbf{54.7} \\
        \midrule
        \multirow{4}{*}{Eyecandies} & w/o IA  & 80.7 & 83.6 & 41.6 & 38.4 \\ 
        ~ & w/o CAE  & 81.9 & 85.4 & 42.4 & 39.3 \\ 
        ~ & w/o $\lambda$ & 82.2 & 85.9 & 43.7 & 40.8 \\
        ~ & Ours & \textbf{82.9} & \textbf{86.1} & \textbf{44.7} & \textbf{41.8} \\
        \bottomrule
    \end{tabular}
    }
\end{table}

\subsubsection{Discussion of the Multimodal Mutual Scoring}
\label{abl:m3sm_module}
Ablation studies in Table \ref{tab:m3sm_module} evaluate four key components: Interval Average (IA), Cross-modal Anomaly Enhancement (CAE), and confidence weight $\lambda$.
Without IA, normal regions with appearance variations receive higher scores from dissimilar patches,
especially in Eyecandies where products have diverse sub-types.
Our IA operation mitigates this issue, improving AP-cls by $2.5\%$ and AP-seg by $3.4\%$.
The CAE module reduces false negatives brought by single-modal invisible anomalies,
boosting MVTec 3D-AD performance by $1.3\%$ F1-max-cls and $2.1\%$ F1-max-seg.
The confidence weight $\lambda$ in our CAE further suppresses cross-modal false positives, enhancing both classification and segmentation.

\begin{table}[!t]
    \centering
    \caption{The ablation of SNAMD module.
    We report the anomaly classification and segmentation results.
    All metrics are in $\%$.}
  \label{tab:snamd_module}
  \setlength{\tabcolsep}{1.0mm}{
    \begin{tabular}{c|cccc}
        \toprule
        Setting & F1-max-cls & AP-cls & F1-max-seg & AP-seg \\ 
        \midrule[0.8pt]
        \multicolumn{5}{c}{MVTec 3D-AD} \\
        \midrule[0.2pt]
        LNA \cite{CVPR2022patchcore} & 92.1 & 94.7 & 43.6 & 39.3 \\ 
        LNA+SWPooling & 92.2 & 96.2 & 53.0 & 52.8 \\ 
        \midrule[0.2pt]
        LNAMD \cite{iclr2024musc} & 92.9 & 95.7 & 47.3 & 44.3 \\ 
        LNAMD+SWPooling & 92.2 & 96.2 & 52.9 & 52.8 \\ 
        \midrule[0.2pt]
        SNAMD (r=1) & 92.2 & 96.1 & 52.8 & 52.7 \\ 
        SNAMD (r=3) & 92.2 & 96.2 & 53.0 & 52.8 \\ 
        SNAMD (r=5) & 92.3 & 96.1 & 53.0 & 52.9 \\ 
        SNAMD (w/o $\exp$) & 92.8 & 96.4 & 50.7 & 48.8 \\
        SNAMD (w/o SWPooling) & \textbf{93.1} & 96.5 & 52.1 & 50.8 \\ 
        SNAMD (Ours) & 93.0 & \textbf{96.8} & \textbf{54.6} & \textbf{54.7} \\
        \midrule[0.8pt]
        \multicolumn{5}{c}{Eyecandies} \\
        \midrule[0.2pt]
        LNA \cite{CVPR2022patchcore} & 75.2 & 77.1 & 33.6 & 29.1 \\ 
        LNA+SWPooling & 81.3 & 84.2 & 44.0 & 40.9 \\ 
        \midrule[0.2pt]
        LNAMD \cite{iclr2024musc} & 77.6 & 78.9 & 36.8 & 33.0 \\ 
        LNAMD+SWPooling & 81.1 & 84.1 & 44.1 & 41.1 \\ 
        \midrule[0.2pt]
        SNAMD (r=1) & 82.3 & 84.8 & 44.4 & 41.4 \\ 
        SNAMD (r=3) & 81.3 & 84.2 & 44.0 & 40.9 \\ 
        SNAMD (r=5) & 80.9 & 83.8 & 43.8 & 40.7 \\
        SNAMD (w/o $\exp$) & 78.2 & 82.1 & 38.4 & 33.8 \\
        SNAMD (w/o SWPooling) & 79.4 & 83.5 & 40.4 & 36.0 \\ 
        SNAMD (Ours) & \textbf{82.9} & \textbf{86.1} & \textbf{44.7} & \textbf{41.8} \\
        \bottomrule
    \end{tabular}
    }
\end{table}

\subsubsection{Discussion of the SNAMD module}
\label{abl:SNAMD_module}
In Table \ref{tab:snamd_module},
we conduct ablation experiments on two main technologies SWPooling and multiple degrees in our SNAMD module.
The results demonstrate that three aggregation degrees $r \in \{1,3,5\}$ outperform using only one aggregation degree.
It brings $0.8\%$ F1-max-cls and $1.6\%$ F1-max-seg improvements on the MVTec 3D-AD dataset,
with consistent gains on Eyecandies.
The $\exp$ operation amplifies the similarity weights of the similar neighborhood patches while suppressing those that are dissimilar.
This effectively reduces the interference of irrelevant backgrounds, leading to consistent performance improvements on both datasets.
Removing SWPooling causes significant performance drops, particularly for the Eyecandies dataset with more small anomalies,
where it reduce false negatives and brings $2.6\%$ AP-cls and $5.8\%$ AP-seg gains.
When integrated into Local Neighborhood Aggregation (LNA) \cite{CVPR2022patchcore} and Local Neighborhood Aggregation with Multiple Degrees (LNAMD) \cite{iclr2024musc}, SWPooling also enhances their performance.

\subsubsection{Effectiveness of our RsCon module}
\label{abl:rscon_module}
In Table \ref{tab:rscin_module}, we perform comprehensive ablation studies of our RsCon module across four datasets.
The consistent improvement in AC metrics across all datasets validates the effectiveness and stability of our RsCon.
Meanwhile, for the window mask operation (WMO) of RsCon, we analyze the window size sensitivity $k \!\in \! \{2,...,9\}$ through box plots in Fig.\ref{fig:param_abl} (a).
Small intervals (Q1$\sim$Q3) across four datasets show stable performance,
which means that RsCon is not sensitive to the window size $k$, except for some extreme values.
Notably, RsCon consistently outperforms baseline methods (red dot) regardless of $k$.
Additionally, removing WMO (black dot) causes significant performance drops, confirming its critical role.

\begin{table}[!t]
    \centering
    \caption{The ablation of the RsCon module across four traditional datasets.
    We report the AC results.
    All metrics are in $\%$.}
  \label{tab:rscin_module}
  \setlength{\tabcolsep}{1.5mm}{
    \begin{tabular}{c|c|ccc}
        \toprule
        Dataset & Setting & AUROC-cls & F1-max-cls & AP-cls \\ 
        \midrule 
        \multirow{2}{*}{MVTec 3D-AD} & w/o RsCon & 86.5 & 92.6 & 96.1 \\ 
        ~ & w RsCon & \textbf{88.1} & \textbf{93.0} & \textbf{96.8} \\
        \midrule
        \multirow{2}{*}{Eyecandies} & w/o RsCon & 83.1 & 82.0 & 85.2 \\ 
        ~ & w RsCon & \textbf{83.9} & \textbf{82.9} & \textbf{86.1} \\
        \midrule
        \multirow{2}{*}{MVTec AD} & w/o RsCon & 94.8 & 94.6 & 97.8 \\ 
        ~ & w RsCon & \textbf{95.3} & \textbf{95.2} & \textbf{98.1} \\
        \midrule
        \multirow{2}{*}{VisA} & w/o RsCon & 86.9 & 84.4 & 87.2 \\ 
        ~ & w RsCon & \textbf{88.2} & \textbf{85.6} & \textbf{88.4} \\
        \bottomrule
    \end{tabular}
    }
\end{table}

\begin{figure}[!t]
\centering
\includegraphics[width=0.49\textwidth]{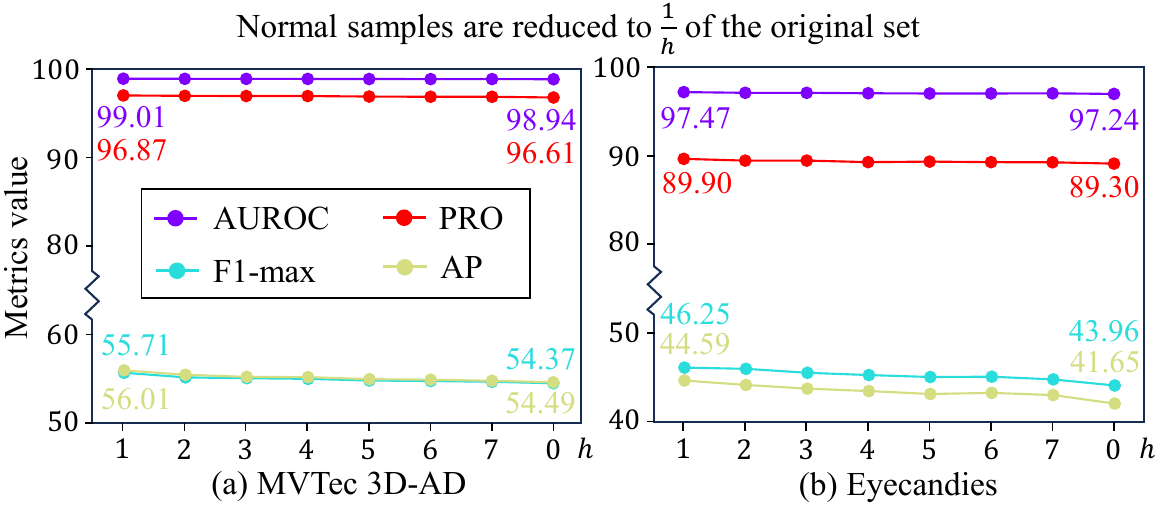}
\caption{
Four anomaly segmentation metrics with different normal sample numbers across MVTec 3D-AD and Eyecandies datasets.
}
\label{fig:normal_abl}
\end{figure}

\subsubsection{Influence of the normal sample number}
\label{abl:normal}
To investigate the robustness of the normal sample number in the test set,
we randomly reduce normal samples to $\frac{1}{h}$ of the original set (Fig. \ref{fig:normal_abl}).
The horizontal axis represents the value of $h$. The limiting case $h=0$ indicates no normal samples in the dataset.
Under this extreme condition,
AS metrics show minimal degradation across both datasets:
AUROC decreases by less than $0.23\%$ and PRO varies by less than $0.6\%$.
The maximum observed drops in F1-max and AP are $2.94\%$, indicating only minor false-positive increases.
These results demonstrate the insensitivity of our MuSc-V2 to normal sample counts and the anomaly percentage,
especially in real industrial scenes where normal samples typically dominate.

\subsubsection{Effect of the dataset size}
\label{abl:subset}
In our mutual scoring mechanism, we use $\{\mathcal{D} \backslash O_i\}$ to assign scores to sample $O_i$.
To explore the sensitivity to dataset size, we divide the unlabeled samples into $g \in \{1,2,3\}$ subsets.
Each subset independently scores samples in this subset.
After averaging results across 10 random seeds, Table \ref{tab:time} shows minimal performance degradation:
AC drops by less than $0.4\%$ and AS declines by less than $1.0\%$.
This demonstrates consistent effectiveness even with limited data.

\subsubsection{Effect of smaller batch size}
\label{abl:batch_size}
In Sec. \ref{abl:subset}, we partition all unlabeled samples into $g \in \{1,2,3\}$ subsets, where mutual scoring is performed only within each subset.
To further investigate the impact of reducing the number of samples that simultaneously participate in mutual scoring, we decrease the batch size to less than 10.
Table \ref{tab:batch_size} reports the results.
When the batch size is extremely small (e.g., 5), our MuSc‑V2 shows limited degradation in AC metrics, with a maximum drop of 1.7\%.
For AS metrics, the performance decreases by 2.4\% on the Eyecandies dataset and 4.3\% on the MVTec 3D‑AD dataset.
This is because, with a batch size of only 5, our interval average operation with a 30\% range involves only one sample in scoring, introducing instability.
Despite this, the overall degradation remains bounded.

\begin{table}[!t]
    \centering
    \caption{The performance of a smaller batch size for mutual scoring.
    We report the AC and AS results. All metrics are in $\%$.}
  \label{tab:batch_size}
  \setlength{\tabcolsep}{1.0mm}{
    \begin{tabular}{c|c|cccc}
        \toprule
        Dataset & Batch size & F1-max-cls & AP-cls & F1-max-seg & AP-seg \\ 
        \midrule
        \multirow{4}{*}{MVTec 3D-AD} & All & 93.0 & 96.8 & 54.6 & 54.7 \\ 
        ~ & 9 & 92.1\scriptsize{($\downarrow$0.9)} & 96.0\scriptsize{($\downarrow$0.8)} & 53.3\scriptsize{($\downarrow$1.3)} & 53.0\scriptsize{($\downarrow$1.7)} \\
        ~ & 7 & 91.9\scriptsize{($\downarrow$1.1)} & 95.8\scriptsize{($\downarrow$1.0)} & 51.7\scriptsize{($\downarrow$2.9)} & 51.4\scriptsize{($\downarrow$3.3)} \\
        ~ & 5 & 91.8\scriptsize{($\downarrow$1.2)} & 95.1\scriptsize{($\downarrow$1.7)} & 50.9\scriptsize{($\downarrow$3.7)} & 50.4\scriptsize{($\downarrow$4.3)} \\
        \midrule
        \multirow{4}{*}{Eyecandies} & All & 82.9 & 86.1 & 44.7 & 41.8 \\ 
        ~ & 9 & 82.8\scriptsize{($\downarrow$0.1)} & 85.9\scriptsize{($\downarrow$0.2)} & 43.9\scriptsize{($\downarrow$0.8)} & 40.5\scriptsize{($\downarrow$1.3)} \\
         ~ & 7 & 82.5\scriptsize{($\downarrow$0.4)} & 85.1\scriptsize{($\downarrow$1.0)} & 43.5\scriptsize{($\downarrow$1.2)} & 40.2\scriptsize{($\downarrow$1.6)} \\
        ~ & 5 & 81.3\scriptsize{($\downarrow$1.6)} & 85.0\scriptsize{($\downarrow$1.1)} & 43.4\scriptsize{($\downarrow$1.3)} & 39.4\scriptsize{($\downarrow$2.4)} \\
        \bottomrule
    \end{tabular}
    }
\end{table}

\begin{figure}[!t]
\centering
\includegraphics[width=0.49\textwidth]{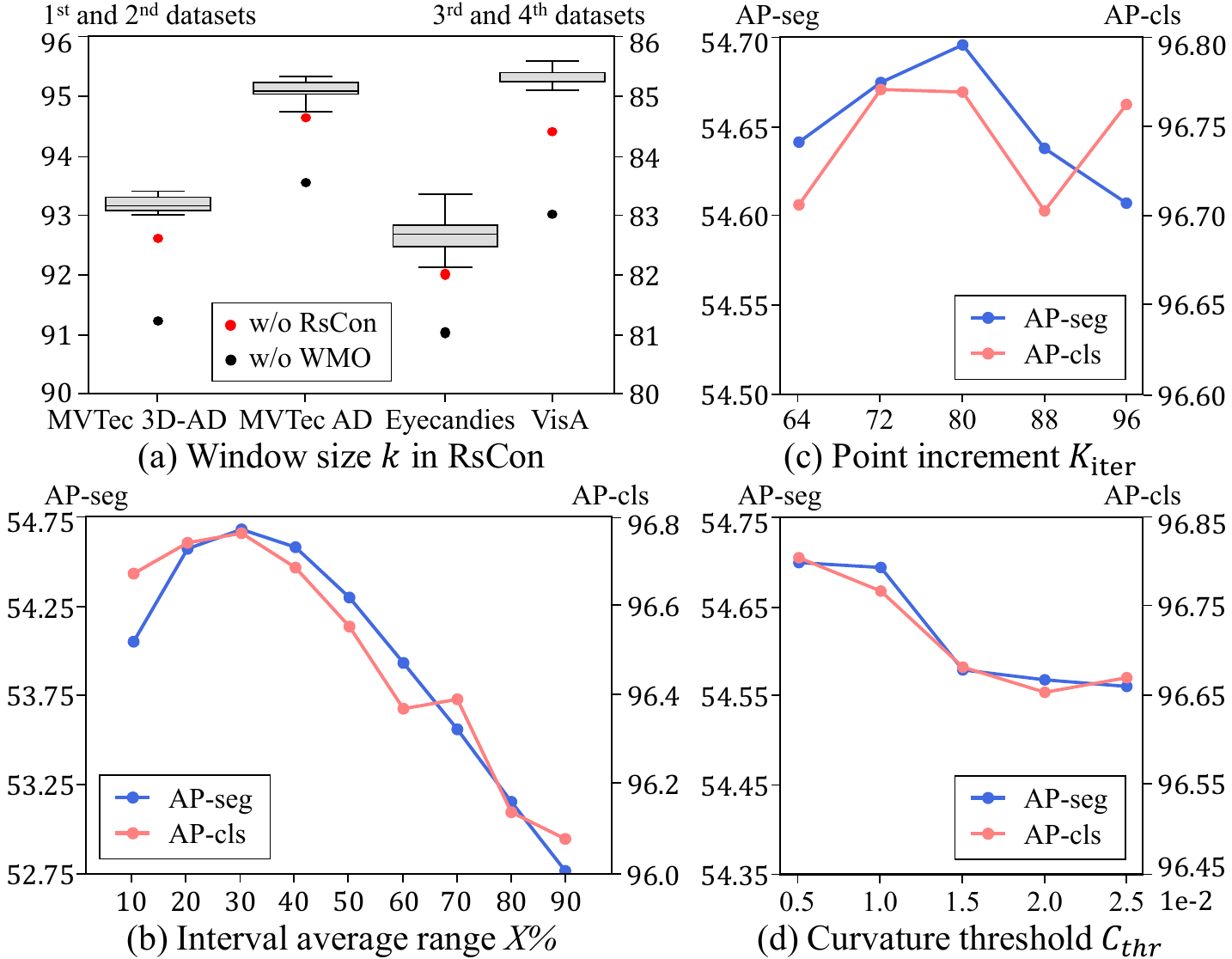}
\caption{
Experimental results of the influence of four hyperparameters on MuSc-V2.
We report AP metrics on MVTec 3D-AD in (b), (c) and (d).
}
\label{fig:param_abl}
\end{figure}

\subsubsection{Sensitivity analysis of hyperparameters}
\label{abl:param}
In Fig. \ref{fig:param_abl}, we conduct experiments on four important hyperparameters in our MuSc-V2.
\textbf{(a)} The RsCon's window size $k$.
This hyperparameter is insensitive as we describe in Sec. \ref{abl:rscon_module}.
\textbf{(b)} The MSM's interval average (IA) range $X\%$.
Larger $X$ values incorporate more normal patches with appearance variations, elevating scores and increasing false positives.
The extreme case $X\!=\!100$ (vanilla average) shows the most severe performance degradation.
While smaller $X$ may bring false negatives since abnormal patches may find a few similar patches.
For an IA range $X\% < 50\%$, the impact remains minimal (AP-cls $\leq$ 0.09\%, AP-seg $\leq$ 0.64\%),
demonstrating our method's robustness to moderate range selections.
\textbf{(c)} The IPG's iterative point increment $K_\text{iter}$.
This parameter demonstrates strong robustness around our default setting ($K_\text{iter}\!=\!80$),
with maximum variations of 0.07\% for AP-seg and 0.09\% for AP-cls.
Performance degrades when $K_\text{iter}$ increases beyond this range, as IPG degradates to traditional KNN.
\textbf{(d)} The IPG's curvature threshold $\mathcal{C}_{thr}$.
This parameter determines which point groups undergo IPG processing.
Higher values reduce the number of processed groups, leaving more groups containing potentially discontinuous surfaces, thus degrading performance.
Near our default setting (0.01), varying $\mathcal{C}_{thr}$ causes minimal impact, where AP-seg and AP-cls fluctuate by less than 1.4\%.
Above analyses demonstrate the robustness of our method's four key hyperparameters within reasonable ranges.
As a training-free approach, maintaining hyperparameters within appropriate bounds is both inevitable and manageable.

\subsubsection{Inference time}
\label{abl:time}
Table \ref{tab:time} compares inference times on an NVIDIA RTX 3090 GPU (excluding I/O) across different zero-shot methods.
Since the MVTec 3D-AD's product range is from 100 to 159, we use 150 samples for mutual scoring in MuSc-V2 and MuSc.
Our MuSc-V2 outperforms its previous version in both accuracy and speed for 2D tasks.
For 3D and multimodal cases, our directly point cloud processing proves significantly faster than PointAD \cite{nips2024pointad},
which requires more than 30s per sample for multi-view rendering.
While subset partitioning further accelerates MuSc-V2, the 722.6ms feature extraction by Point-MAE \cite{eccv2022pointmae} remains a bottleneck.

\begin{table}[!t]
    \centering
    \caption{Per sample inference time of our MuSc-V2 and other zero-shot methods.
    We divide the MVTec 3D-AD dataset into $g$ subsets.}
  \label{tab:time}
  \setlength{\tabcolsep}{1.5mm}{
    \begin{tabular}{c|c|cc|c}
        \toprule
        Method & Train & F1-max-cls & F1-max-seg & Time (ms) \\ 
        \midrule[0.8pt]
        \multicolumn{5}{c}{2D modal (ViT-L-14-336)} \\
        \midrule[0.2pt]
        AdaCLIP\cite{eccv2024adaclip} & \checkmark & 89.9 & 41.7 & 301.4 \\
        VCP-CLIP\cite{eccv2024vcpclip} & \checkmark & 89.9 & 42.5 & 86.2\\
        MuSc\cite{iclr2024musc} &  & \textbf{90.3} & 41.0 & 737.1 \\
        MuSc-V2($g$=1) &  & \textbf{90.3} & \textbf{47.1} & 122.3 \\
        MuSc-V2($g$=2) &  & 90.3 ($\downarrow$0.0) & 46.8 ($\downarrow$0.3) & 92.9 \\
        MuSc-V2($g$=3) &  & 90.0 ($\downarrow$0.3) & 46.9 ($\downarrow$0.2) & 85.6 \\
        \midrule[0.8pt]
        \multicolumn{5}{c}{2D modal (ViT-B-8)} \\
        \midrule[0.2pt]
        MuSc &  & 90.0 & 31.9 & 245.9 \\
        MuSc-V2($g$=1) &  & \textbf{90.5} & \textbf{40.4} & 44.1 \\
        MuSc-V2($g$=2) &  & 90.3 ($\downarrow$0.2) & 40.1 ($\downarrow$0.3) & 35.4 \\
        MuSc-V2($g$=3) &  & 90.2 ($\downarrow$0.3) & 39.7 ($\downarrow$0.7) & 32.7 \\
        \midrule[0.8pt]
        \multicolumn{5}{c}{3D modal} \\
        \midrule[0.2pt]
        PointAD\cite{nips2024pointad} & \checkmark & 92.3 & 30.7 & 337.1+30287.7 \\
        MuSc-V2($g$=1) &  & \textbf{92.5} & \textbf{45.9} & 745.8 \\
        MuSc-V2($g$=2) &  & 92.3 ($\downarrow$0.2) & 45.6 ($\downarrow$0.3) & 737.6 \\
        MuSc-V2($g$=3) &  & 92.2 ($\downarrow$0.1) & 45.4 ($\downarrow$0.5) & 733.9 \\
        \midrule[0.8pt]
        \multicolumn{5}{c}{Multimodal} \\
        \midrule[0.2pt]
         PointAD\cite{nips2024pointad} & \checkmark & 92.2 & 37.2 & 368.2+30287.7\\
        MuSc-V2($g$=1) &  & \textbf{93.0} & \textbf{54.6} & 969.6 \\
        MuSc-V2($g$=2) &  & 92.7 ($\downarrow$0.3) & 54.3 ($\downarrow$0.3) & 943.7 \\
        MuSc-V2($g$=3) &  & 92.5 ($\downarrow$0.4) & 53.6 ($\downarrow$1.0) & 934.4 \\
        \bottomrule
    \end{tabular}
    }
\end{table}

\section{Conclusion}
In this paper, we present MuSc-V2, a zero-shot framework for industrial anomaly classification and segmentation in multimodal data.
This method leverages implicit normal/abnormal cues from the unlabeled samples.
We propose four key innovations:
(1) SNAMD modules for modeling anomalies with varying scales;
(2) IPG modules for generating 3D groups with continuous surfaces and maintaining the normal representation consistency;
(3) a multimodal mutual scoring mechanism for scoring each sample patch;
(4) RsCon for false classifications suppression.
Experimental results demonstrate superior performance over existing zero-shot methods, with competitive advantages against few-shot approaches.

\bibliographystyle{IEEEtran}
\bibliography{main}

\begin{IEEEbiography}[{\includegraphics[width=1in,height=1.25in,clip,keepaspectratio]{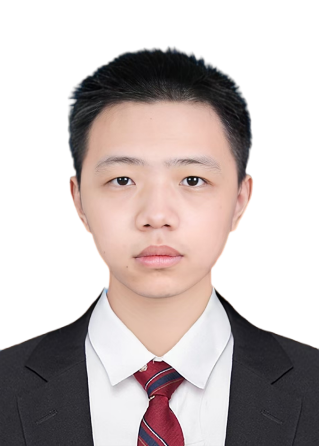}}]{Xurui Li}
received the B.S. degree in electronics and information engineering from the Huazhong University of Science and Technology (HUST), Wuhan, China, in 2023. He is currently pursuing the Ph.D. degree with the Huazhong University of Science and Technology. His research interests include computer vision and anomaly detection.
\end{IEEEbiography}

\begin{IEEEbiography}[{\includegraphics[width=1in,height=1.25in,clip,keepaspectratio]{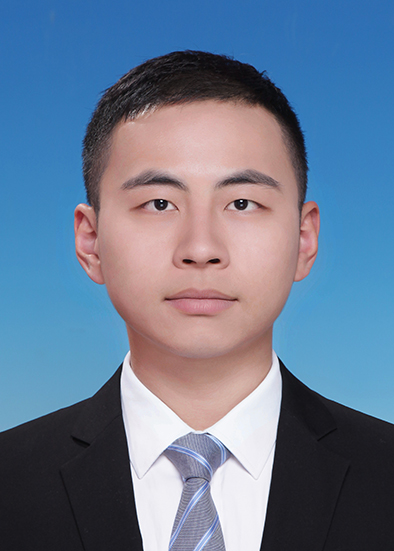}}]{Feng Xue}
 (Member, IEEE) received the Ph.D. degree from the Beijing University of Posts and Telecommunications (BUPT), Beijing, China, in 2023. He is currently a Post-Doctoral Researcher with the University of Trento. His research interests include open-world visual perception and its applications in various fields, such as road segmentation, monocular depth estimation, and object detection.
\end{IEEEbiography}

\begin{IEEEbiography}[{\includegraphics[width=1in,height=1.25in,clip,keepaspectratio]{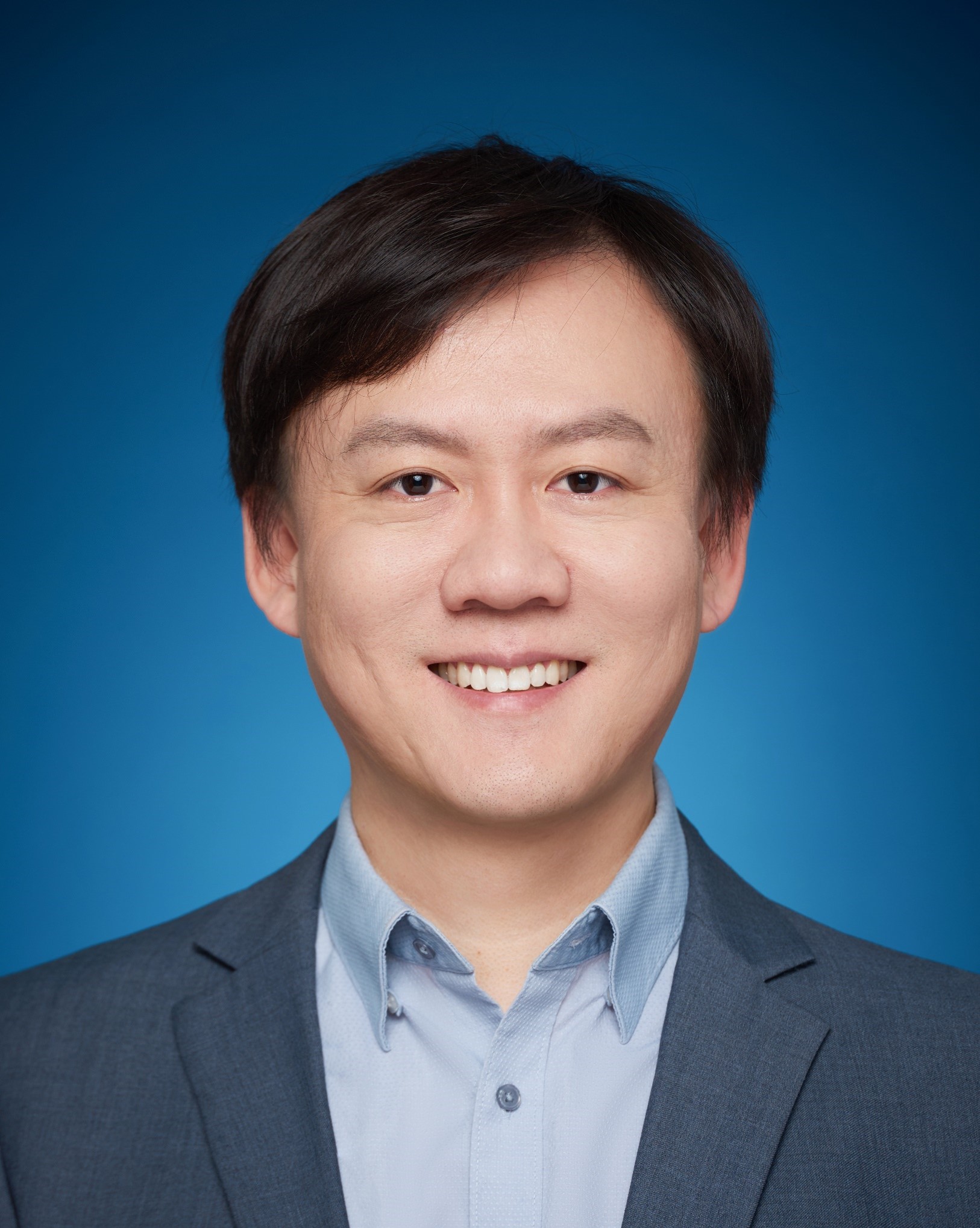}}]{Yu Zhou}
(Member, IEEE) received the M.S. and Ph.D. degrees in electronics and information engineering from the Huazhong University of Science and Technology (HUST), Wuhan, China in 2010, and 2014, respectively. In 2014, he joined the Beijing University of Posts and Telecommunications (BUPT), Beijing, as a Post-Doctoral Researcher from 2014 to 2016 and an Assistant Professor from 2016 to 2018. He is currently a Professor with the School of Electronic Information and Communications, HUST. His research interests include computer vision and industrial vision.
\end{IEEEbiography}

\end{document}